\documentclass[sigconf]{acmart}

\AtBeginDocument{%
  \providecommand\BibTeX{{%
    \normalfont B\kern-0.5em{\scshape i\kern-0.25em b}\kern-0.8em\TeX}}}

\setcopyright{acmcopyright}
\copyrightyear{2025}
\acmYear{2025}
\acmDOI{10.1145/3711896.3736854}
\acmConference[KDD '25]{The 31st  ACM SIGKDD Conference on Knowledge Discovery and Data Mining}{June 03--05,
  2018}{Woodstock, NY}
\acmPrice{15.00}
\acmISBN{978-1-4503-XXXX-X/18/06}

\usepackage{balance}
\usepackage{booktabs}
\usepackage{multirow}
\usepackage{graphicx}
\usepackage{amsmath}
\usepackage{subcaption}
\usepackage{pgfplots}
\usepackage{wrapfig}
\usepackage{adjustbox}
\usepackage{array}
\usepackage{bm}
\usepackage{enumitem}
\usepackage{colortbl}

\newtheorem{definition}{Definition}
\newtheorem*{myproblem}{Problem}
\definecolor{customcolor}{HTML}{ECF4FF}

\copyrightyear{2025}
\acmYear{2025}
\setcopyright{acmlicensed}\acmConference[KDD '25]{Proceedings of the 31st ACM SIGKDD Conference on Knowledge Discovery and Data Mining V.2}{August 3--7, 2025}{Toronto, ON, Canada}
\acmBooktitle{Proceedings of the 31st ACM SIGKDD Conference on Knowledge Discovery and Data Mining V.2 (KDD '25), August 3--7, 2025, Toronto, ON, Canada}
\acmDOI{10.1145/3711896.3736854}
\acmISBN{979-8-4007-1454-2/2025/08}

\begin{document}
\title{Beyond Fixed Variables: Expanding-variate Time Series Forecasting via Flat Scheme and Spatio-temporal Focal Learning}

\author{Minbo Ma}
\affiliation{
    \institution{Southwest Jiaotong University}
    \city{Chengdu}
    \country{China}
}
\authornote{School of Computing and Artificial Intelligence, Southwest Jiaotong University}
\affiliation{
    \institution{FernUniversität in Hagen}
    \city{Hagen}
    \country{Germany}
}
\email{minboma@my.swjtu.edu.cn}

\author{Kai Tang}
\authornotemark[1]
\affiliation{
    \institution{Southwest Jiaotong University}
    \city{Chengdu}
    \country{China}
}
\email{tangkailh@outlook.com} 

\author{
    Huan Li
}
\affiliation{
    \institution{College of Computer Science and Technology}
    \institution{Zhejiang University}
    \city{Hangzhou}
    \country{China}
}
\email{lihuan.cs@zju.edu.cn}

\author{Fei Teng}
\authornotemark[1]
\affiliation{
    \institution{Southwest Jiaotong University}
    \city{Chengdu}
    \country{China}
}
\authornote{Fei Teng and Dalin Zhang are the corresponding authors.}
\affiliation{
    \institution{Engineering Research Center of Sustainable Urban Intelligent Transportation}
    \city{Chengdu}
    \country{China}
}
\email{fteng@swjtu.edu.cn}

\author{
    Dalin Zhang
}
\authornotemark[2]
\affiliation{
    \institution{Department of Computer Science}\institution{Aalborg University}
    \city{Aalborg}
    \country{Denmark}
}
\email{dalinz@cs.aau.dk}

\author{
    Tianrui Li
}
\authornotemark[1]
\affiliation{
    \institution{Southwest Jiaotong University}
    \city{Chengdu}
    \country{China}
}
\affiliation{
    \institution{Engineering Research Center of Sustainable Urban Intelligent Transportation}
    \city{Chengdu}
    \country{China}
}
\email{trli@swjtu.edu.cn}

\begin{abstract}
Multivariate Time Series Forecasting (MTSF) has long been a key research focus. Traditionally, these studies assume a fixed number of variables, but in real-world applications, Cyber-Physical Systems often expand as new sensors are deployed, increasing variables in MTSF. 
In light of this, we introduce a novel task, \textit{Expanding-variate Time Series Forecasting}  (EVTSF). 
This task presents unique challenges, specifically
(1) handling inconsistent data shapes caused by adding new variables, and (2) addressing imbalanced spatio-temporal learning, where expanding variables have limited observed data due to the necessity for timely operation. To address these challenges, we propose STEV, a flexible spatio-temporal forecasting framework. 
STEV includes a new \textit{Flat Scheme} to tackle the inconsistent data shape issue, which extends the graph-based spatio-temporal modeling architecture into 1D space by flattening the 2D samples along the variable dimension, making the model variable-scale-agnostic while still preserving dynamic spatial correlations through a holistic graph.
Additionally, we introduce a novel \textit{Spatio-temporal Focal Learning} strategy that incorporates a negative filter to resolve potential conflicts between contrastive learning and graph representation, and a focal contrastive loss as its core to guide the framework to focus on optimizing the expanding variables. 
To evaluate the effectiveness of STEV, we benchmark EVTSF performance on three real-world datasets from various domains and compare it against three potential solutions employing state-of-the-art (SOTA) MTSF models tailored for EVSTF. 
Experimental results show that STEV significantly outperforms its competitors, especially in handling expanding variables.
Notably, STEV, with only 5\% of observations during the expanding period, is on par with SOTA MTSF models trained with complete data. 
Further exploration of various expanding scenarios underscores the generalizability of STEV in real-world applications.
\end{abstract}


\maketitle

\begin{figure}[!ht]
    \centering
    \includegraphics[width=0.9\columnwidth]{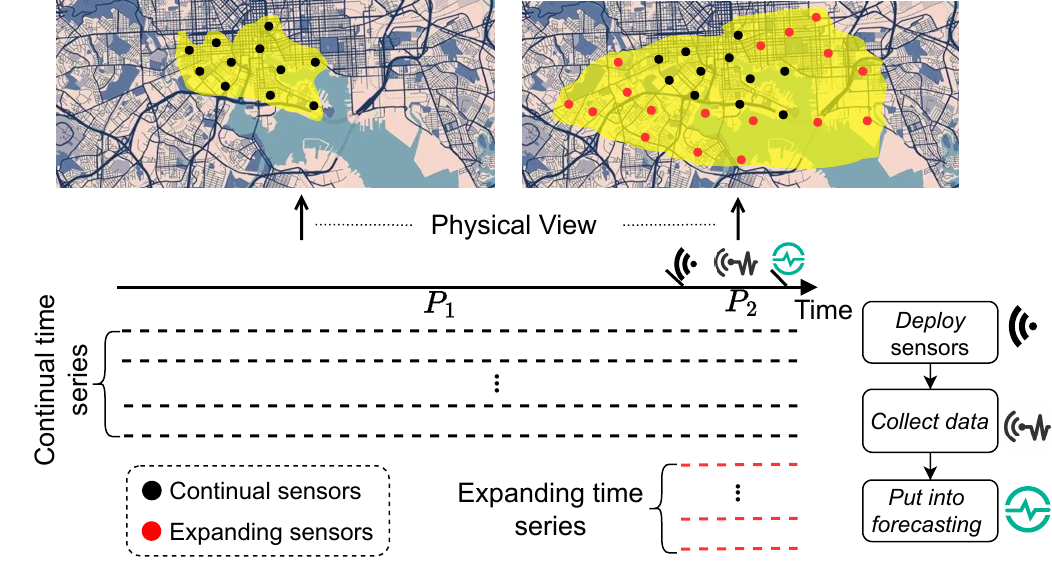}
    \caption{Expanding-variate Time Series in traffic monitoring. Left: Initial system with continual sensors (black dots) during period $P_1$. Right: Expanded system with both continual and expanding sensors (black and red dots) during $P_2$.}
    \label{fig:background}
\end{figure}
\setlength{\textfloatsep}{10pt}

\section{Introduction}
Multivariate Time Series (MTS), reflecting the fluctuation of multiple sensors' measurements over time, plays a vital role in understanding and optimizing Cyber-Physical Systems (CPS) across various domains~\cite{bai2020adaptive,zheng2013u,wang2023Cons}. As a fundamental task and a research hotspot in MTS utilization, MTS Forecasting (MTSF) necessitates a keen consideration of both the temporal dynamics of each variable and the spatial correlations across variables~\cite{Li2018DCRNN, NEURIPS2023_dc1e32dd, liu2023itransformer}. 
While recent advancements in MTSF have focused on developing novel deep learning networks to enhance forecasting accuracy, progress has nearly plateaued, often neglecting critical real-world requirements.

In this study, we instead draw attention to a frequently overlooked scenario in MTS research: \emph{Expanding-variate Time Series} (EVTS), where the number of variables dynamically increases over time. Specifically, we explore situations where an established CPS evolves with newly deployed sensors, which are expected to function with their supporting model as soon as possible.
Consider a traffic management system, as illustrated in Fig.~\ref{fig:background}. Initially, sensors (black dots) monitor traffic conditions within a designated area (yellow shadow). As urban development progresses, additional sensors (red dots) are deployed, expanding the monitoring area (expanded yellow shadow). In this evolving context, we define the initially deployed sensors as \emph{continual sensors}, and their measurements as \emph{continual variables}, while the newly deployed sensors and their measurements are termed \emph{expanding sensors} and \emph{expanding variables}, respectively.

We will focus on the EVTS Forecasting (EVTSF) task, which aims to accurately forecast future values of both continual and expanding variables. In particular, the newly installed sensors are expected to perform forecasting upon deployment. 
Early forecasting is highly beneficial. For instance, in transportation, timely predictions of future traffic conditions in newly monitored regions can help reduce accidents during critical early stages. 
Similarly, in power systems, substantial ice accumulation on transmission lines can cause widespread power outages, as has been extensively documented in both China~\cite{example1} and the United States~\cite{example2}. To mitigate such risks, monitoring devices like tension sensors are deployed to detect ice buildup, where tension forecasting supports prompt de-icing measures. In both scenarios, initiating forecasts immediately after sensor deployment is essential for effective risk prevention and mitigation. 

Despite its substantial advantages, this task not only shares the common challenge of capturing spatio-temporal dependencies but also faces two additional major challenges:
\begin{itemize}[leftmargin=*]
    \item \textbf{Challenge 1: Inconsistent data shapes.} As new variables are introduced, the shape of the data changes, as shown in Fig.~\ref{fig:background}. This necessitates that deep model training adapts to varying dimensions within the mini-batch, a stark contrast to the equal-sized data shapes assumed by traditional training methods. An intuitive strategy is to treat EVTS as a set of independent univariate time series (UTS) and perform forecasting for each UTS individually. However, this approach fails to leverage correlations among variables, limiting its effectiveness. Another strategy is to apply the padding operation to standardize sample shapes, but this often introduces noise, resulting in inevitably inferior performance (refer to the evaluation in Section~\ref{sec:overall}). Besides, several studies~\cite{chen2021trafficstream, wang2023pattern} employ continual learning~\cite{continuallearning}, which sequentially updates models using newly available data. While this approach is feasible in dynamic settings, its effectiveness is constrained by the catastrophic forgetting problem~\cite{kirkpatrick2017overcoming}, i.e., neural models significantly forget previously learned information upon learning new data.
    
    \item \textbf{Challenge 2: Imbalanced spatio-temporal learning}. Newly deployed sensors need to start providing accurate forecasts as soon as possible. However, collecting enough data to fully capture the new spatio-temporal patterns takes time. This leads to a significant imbalance in the volume of data available, with the long-term observations ($P_1+P_2$ in Fig.~\ref{fig:background}) of continual variables and only very short-term data ($P_2$ in Fig.~\ref{fig:background}) of expanding variables. As a result, models trained under these conditions are often biased toward the well-represented continual variables, resulting in less accurate predictions for the newly added expanding variables. Conventional methods focus on increasing samples of expanding variables through techniques like oversampling and data augmentation. However, these approaches can easily lead to overfitting, especially when the synthetic data fails to adequately represent the expanding variables. Besides, previous studies have employed large language model fine-tuning~\cite{gpt4ts} and channel-independent approaches~\cite{PatchTST} to enhance models with few-shot learning capabilities, they face challenges in capturing sequential dependencies in time series and effectively modeling spatial dependencies~\cite{tan2024language}.
\end{itemize}

\begin{figure*}[!htbp]
    \centering
    \includegraphics[width=0.6\linewidth]{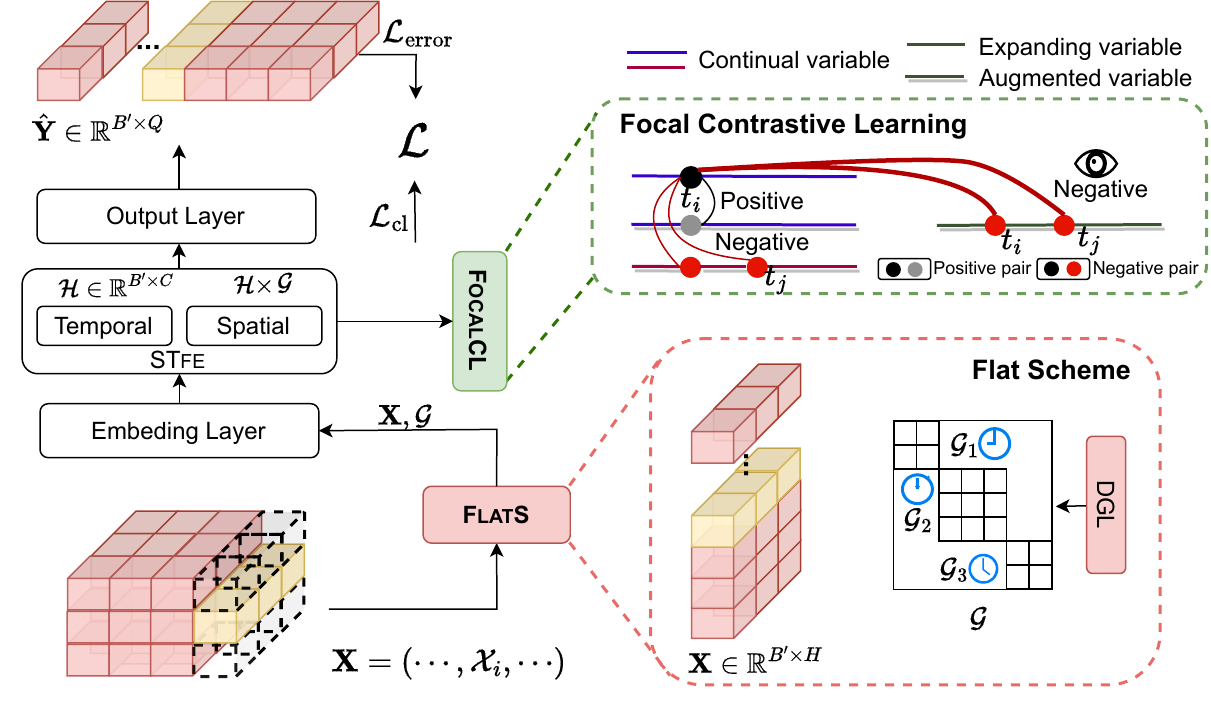}
    \caption{The overview of the proposed framework STEV for EVTSF. Given EVTS data, STEV includes two components to overcome the inconsistent shape and imbalanced learning issues. Within the \textit{Flat Scheme}, EVTS data are flattened along the variable dimension. Meanwhile, DGL generates time-aware graph adjacency matrices. \textit{Focal Contrastive Learning} aims to learn discriminative representations from outputs of spatio-temporal feature extractor (\textsc{STfe}), especially for expanding variables through focal contrastive loss. The joint loss combines the forecasting error loss and contrastive loss.}
    \label{fig:framework}
% \vspace{-13pt}
\end{figure*}
To tackle these challenges, we propose STEV~\footnote{\url{https://github.com/mb-Ma/STEV}}, a generic \underline{S}patio-\underline{T}emporal neural forecasting framework for \underline{EV}TSF.
For \textbf{Challenge 1}, we introduce a simple yet effective \textit{Flat Scheme} (\textsc{FlatS}), which flattens EVTS along the variable dimension, transforming it into UTS, thereby making the model variable scale-agnostic. To preserve the spatial correlations in the original EVTS, we then present a holistic graph where isolated subgraphs maintain these spatial relationships. 
% Each subgraph adapts to learnable, time-aware spatial correlations through a dynamic graph learning scheme.
For \textbf{Challenge 2}, inspired by the success of contrastive learning in capturing powerful representations from limited data~\cite{chen2020simple, yue2022ts2vec, NEURIPS2023_48aaa5ea}, we introduce \textit{Spatio-temporal Focal Learning} (STFL) to guide the model in learning more effective representations for the undersampled expanding variables.
Our approach begins with negative sample filtering, designed to resolve conflicts between the negative sample generation strategy and the spatial neighbouring similarities. 
Building on this, we then introduce \textit{Focal Contrastive Learning} (\textsc{FocalCL}), which incorporates a newly designed focal temperature factor into the contrastive learning loss, ensuring that the optimization process prioritizes expanding variables, thereby enhancing learning effectiveness.
Extensive experiments on three real-world EVTS datasets confirm that STEV not only outperforms potential state-of-the-art (SOTA) solutions but also demonstrates robustness across various expanding scenarios. In summary, our contributions are fourfold:
\begin{enumerate}[leftmargin=*]
    \item We introduce an emerging task, EVTSF, addressing a frequently overlooked aspect of CPS, that is evolving with sensing expansion. In response, we present STEV, a pioneering framework designed to tackle the unique challenges of EVTSF. (Sections~\ref{pd} \& \ref{sec:method})
    \item We develop \textsc{FlatS}, a solution to the challenge of inconsistent-shaped mini-batches specific to EVTSF. This straightforward yet effective method preserves original spatial correlations while transforming variable-shaped data into a consistent format. (Section~\ref{sec:flat})
    \item To address the issue of imbalanced spatio-temporal learning, we propose STFL, an unsupervised auxiliary task that enhances the learning of spatio-temporal representations for the undersampled expanding variables. (Section~\ref{sec:contrastive})
    \item Our experiments involve creating three diverse EVTS datasets across various domains to evaluate the STEV framework. The results show that STEV delivers performance comparable to SOTA MTSF methods using complete data, even when leveraging only 5\% of the observed data. (Section~\ref{sec:exp})
\end{enumerate}

\section{Problem Definition}~\label{pd}
We focus on defining the EVTSF problem within an individual expansion process. Consecutive expansions can be treated as repeated instances of this process. 
\begin{definition}[Expanding-variate Time Series]
Let $\mathcal{V}_1$ and $\mathcal{V}_2$ denote the sets of variables before and after expansion, respectively. $\mathcal{V}_1$ includes only continual variables, while $\mathcal{V}_2$ encompasses both continual and expanding variables, thus $\mathcal{V}_1 \subset \mathcal{V}_2$.
The EVTS dataset $\mathcal{D}$ consists of two main components: MTS data before variable expansion $\mathcal{D}_1 \in \mathbb{R}^{|\mathcal{V}_1| \times |P_1|}$ and after variable expansion $\mathcal{D}_2 \in \mathbb{R}^{|\mathcal{V}_2| \times |P_2|}$, i.e., \(\mathcal{D}=(\mathcal{D}_1, \mathcal{D}_2)\). 
To ensure seamless forecasting upon new sensor installation, the observed time steps after expansion should be significantly fewer than those before expansion, i.e., $|P_2| \ll |P_1|$. 
\end{definition}

\begin{definition}[Expanding-variate Time Series Forecasting]
We aim to build an EVTSF function $\mathcal{F}$ that forecasts the future values for both continual and expanding variables. Formally,
\begin{equation}
    \hat{\mathbf{Y}} = \mathcal{F}(\mathbf{X}),
\end{equation}
where $\mathbf{X} \in \mathbb{R}^{|\mathcal{V}_2| \times H}$ represents the historical observations and $\hat{\mathbf{Y}} \in \mathbb{R}^{|\mathcal{V}_2| \times Q}$ represents the predictions, $H$ and $Q$ denote the historical and forecasting horizons, respectively.
\end{definition}

\begin{myproblem}[Expanding-variate Time Series Forecasting]
We define the data-driven optimization problem for $\mathcal{F}^*(\cdot)$ with the parameter $\theta$ as follows:
\begin{equation}\label{eq:opt}
    \mathcal{F}_\theta^* \gets \underset{\mathcal{\mathcal{F}_\theta}}{\text{argmin}} \; \mathbb{E}_{\mathcal{X}_{i}, \mathcal{Y}_{i} \sim \mathcal{D}} \left[ \mathcal{L}(\mathcal{F}_\theta(\mathcal{X}_{i}), \mathcal{Y}_{i}) \right],
\end{equation}
where $\mathcal{X}_i \in \mathbb{R}^{\mathcal{N} \times H}, \mathcal{Y}_i\in \mathbb{R}^{\mathcal{N} \times Q}$ are drawn from $\mathcal{D}$ using the sliding window technique, and $\mathcal{L}(\cdot)$ is the objective loss function. Notably, we mix and shuffle the data from before and after the variable expansion during the model optimization process. Therefore, the variable dimension $\mathcal{N}$ in Equation (\ref{eq:opt}) equals to $|\mathcal{V}_1|$ if the samples are from before expansion, and $|\mathcal{V}_2|$ otherwise.
\end{myproblem}
 
\section{Method}~\label{sec:method}
\vspace{-20pt}
\subsection{Overview}
Figure~\ref{fig:framework} provides an overview of STEV, highlighting two innovations integrated into general MTSF models: \textsc{FlatS} and \textsc{FocalCL}.
As discussed in Challenge 1, EVTS data contains varying numbers of variables, whereas conventional methods generally handle regular tensors. 
To address this issue, we propose \textit{\textsc{FlatS}} to flatten the EVTS along the variable dimension, enabling a model to handle an arbitrary number of variables. 
We additionally present a \textit{holistic isolated graph} to preserve the spatial correlations within variables, which are disordered when flattening.
For each subgraph within the holistic isolated graph, we introduce the \textit{dynamic graph learning} (DGL) scheme to adaptively represent the time-dependent spatial correlations.  
Addressing Challenge 2, we draw inspiration from contrastive learning's ability to derive robust representations from limited labeled data. We propose \textit{\textsc{STFL}} with a negative sample filter and a \textit{\textsc{FocalCL}} to derive robust and discriminative representations. 
%\textit{Spatio-temporal Focal Learning} (\textsc{STFL}) \textit{Focal Contrastive Learning} (\textsc{FocalCL})
These two custom mechanisms enable the model to avoid hard negative samples and focus on optimizing representations of expanding variables with limited observed data. The final optimization objective $\mathcal{L}$ is a combination of traditional forecasting loss $\mathcal{L}_{\text{error}}$ and \textsc{FocalCL} loss $\mathcal{L}_{\text{cl}}$. 
\vspace{-10pt}
\subsection{Flat Scheme (\textsc{FlatS})}\label{sec:flat}
The variable expansion causes inconsistencies in sample data shapes within a mini-batch, rendering traditional MTSF training methods that depend on uniform shapes ineffective for the EVTSF task. 
One straightforward solution to this issue is the padding scheme, where samples are padded with constant values (e.g., zeros) to achieve uniformity, and a graph is to represent inter-variable correlations, as shown in the upper part of Figure~\ref{fig:trainingscheme}. 
However, this approach inevitably introduces disturbances from the padding values, which persist despite applying a mask to the loss function to minimize their impact. To address this problem, we propose a simple yet effective \textsc{FlatS}. As shown in the bottom part of Figure~\ref{fig:trainingscheme}, we first transform 2D MTS data into 1D univariate time series (UTS) data and then employ a holistic isolated graph to retain the spatial correlations among the flat univariate time series.

\subsubsection{Flat Transformation}
\begin{figure}[!ht]
    \centering
    \includegraphics[width=\columnwidth]{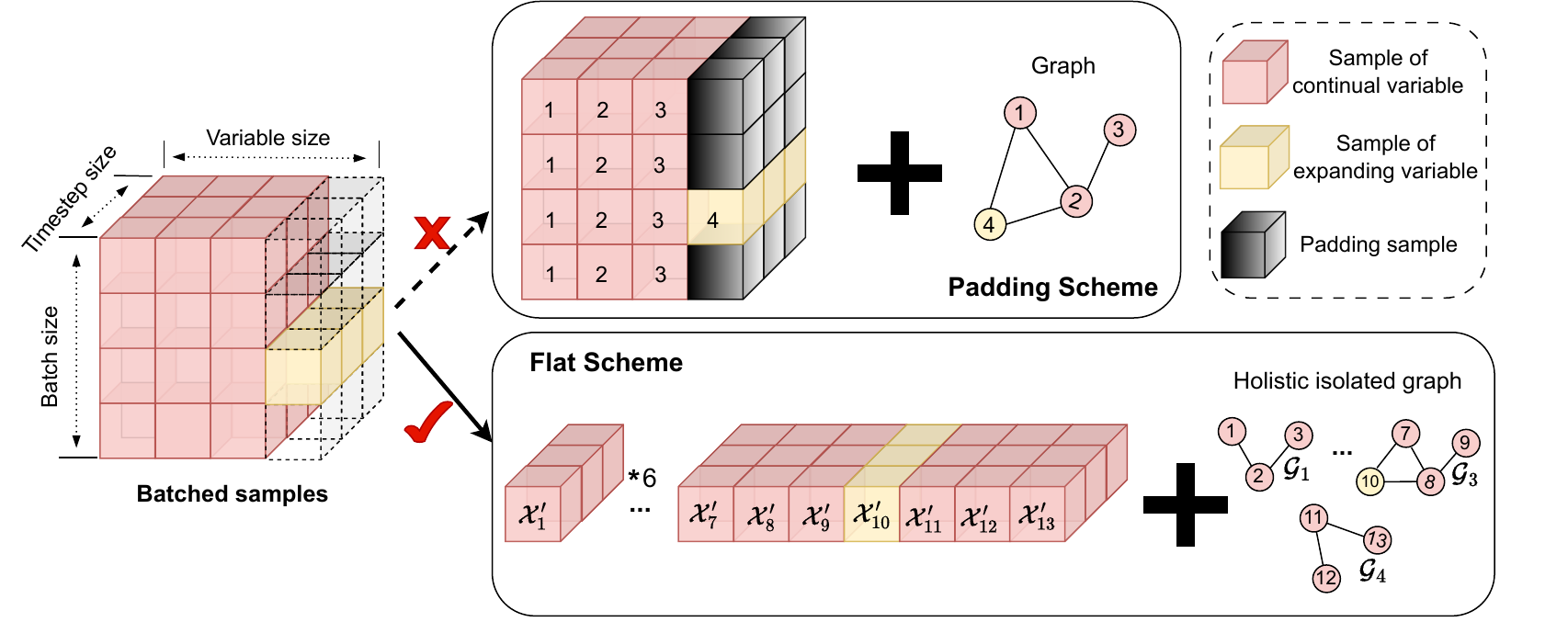}
    \caption{\textsc{FlatS} vs. padding scheme for EVTSF.}
    \label{fig:trainingscheme}
% \vspace{-10pt}
\end{figure}

Suppose we have a mini-batch EVTS samples \(\mathbf{X}=\{\mathcal{X}_i\}_{i=1}^{B}\) (Figure~\ref{fig:trainingscheme} left), where \(B\) is the batch size and $\mathcal{X}_i$ is an EVTS sample.
\(\mathcal{X}_i \in \mathbb{R}^{|\mathcal{V}_1| \times H}\) if it is before expansion and \(\mathcal{X}_i \in \mathbb{R}^{|\mathcal{V}_2| \times H}\) if it is after expansion.
To handle the inconsistent data shapes, we flatten \(\mathbf{X}\) along the variable dimension, transforming it into a new mini-batch $\mathbf{X'}=[\mathcal{X'}_i|i=1,2, ..., B']$ (Figure~\ref{fig:trainingscheme} bottom), where $\mathcal{X'}_i\in \mathbb{R}^H$ is an UTS and $B'\in[B\times|\mathcal{V}_1|, B\times|\mathcal{V}_2|]$ represents the total count of samples in the mini-batch \(\mathbf{X'}\). This flattening process converts the original batch $\mathbf{X}$ with different-shaped MTS samples into a new batch $\mathbf{X'}$ with uniform-shaped UTS samples, ensuring consistency in data shape for effective model training. However, this process loses the correlations among variables in the original samples. Therefore, we propose a holistic isolated graph to preserve the essential relationships between variables, enabling seamless spatial feature extraction. 

\subsubsection{Holistic Isolated Graph} 
The original EVTS $\mathbf{X}$ uses a graph structure to represent spatial correlations among variables. Inspired by the advanced mini-batching module of PyG~\cite{fey2019fast}, which efficiently handles multiple graphs with varying numbers of nodes by stacking them diagonally, we propose the holistic isolated graph to maintain these spatial correlations within the flattened samples.

In the original graph (Figure~\ref{fig:trainingscheme} top), each sample shares a uniform structure. Differently, the holistic isolated graph (Figure~\ref{fig:trainingscheme} bottom) comprises a disconnected collection of $B$ distinct subgraphs $\{\mathcal{G}_i\}_{i=1}^B$, each corresponding to consecutive UTS samples in $\mathbf{X'}$, which align with their pre-expanding sample $\mathcal{X}_i$ in $\mathbf{X}$. For instance, in the mini-batch in Figure~\ref{fig:trainingscheme}, before flattening, the original graph represents the spatial correlations of the post-expansion sample that is $\mathcal{X}_3$, containing variables 1 to 4. After flattening, samples $\mathcal{X'}_7$ to $\mathcal{X'}_{10}$ correspond to the original sample $\mathcal{X}_3$ and share a subgraph $\mathcal{G}_3$, while the other flat samples have their own subgraphs. This method not only ensures the spatial relationships are retained but also keeps a specific subgraph for each original sample.

% Although the form of univariate time series through flat transform is immune to variable changes, the graph structure needs to be reorganized to accommodate the flat samples. 
% %Before introducing the holistic isolated graph, we briefly present how to construct the graph by dynamic graph learning module.
% We first introduce the effect of the graph structure. 
% In graph-based MTSF methods, variables and correlations among variables are treated as nodes and edges, and time series are seen as node features.
% Based on the structure, features between nodes can be aggregated to learn complex spatial representations.
% Consequently, after transformation, the corresponding indexes between nodes and node features need to be reconstructed.

%The graph is a disconnected graph consisting of $B$ distinct subgraphs. Taking the batched samples in Figure~\ref{fig:trainingscheme} as an example, $\mathcal{G}_i$ denotes the graph of $b_i$ flat samples, where an expanding variable is in $3$-th sample. 
% We reassign the node index to ensure that node features aggregate information from their neighbors, e.g., $4$-th node changes to $10$-th node.

Then we introduce the DGL module to construct the subgraph $\mathcal{G}_i$. Considering that in real-world CPS, pre-defined spatial correlations are not always available, we propose to construct learnable graph structures. In addition, the spatial correlations are not static but reveal periodic characteristics as time evolves due to regular production activities~\cite{ma2023learning}. 
The adjacency matrix of the learnable time-aware subgraph $\mathcal{G}_i$ at time $t$ is derived as follows:
\begin{align}
    \mathbf{A}_{t,i} &= \textsc{Sparse}(\mathbf{E}_t\cdot \mathbf{E}_t^\top),\; \text{where}\;\mathbf{E}_t = \textsc{Cat}(\mathbf{E}, e_t).
\end{align}
Here, $\mathbf{E}\in \mathbb{R}^{N \times d}$ represents the $d$-dimensional learnable embeddings of the $\mathcal{N}$ variables, and $e_t=\textsc{Cat}(e_\text{tod}, e_\text{dow})$ is a joint periodic embedding combining the time-of-day (tod) and the day-of-week (dow). The $\textsc{Cat}(\cdot,\cdot)$ operation concatenates these embeddings concatenation operation. The periodic embedding is also learnable during training and indexable during inference. 
The $\textsc{Sparse}(\cdot)$ function applies a sigmoid activation followed by a filter operation that retains positive correlations, enhancing generalization while reducing computational overhead. This \textsc{dgl} scheme ensures that the spatial correlations within the subgraph $\mathcal{G}_i$ are both adaptive and reflective of temporal periodicities.

\subsection{Spatio-temporal Focal Learning (STFL)}~\label{sec:contrastive}
After flattening the original mini-batch \(\mathbf{X}\), we use a spatio-temporal feature extractor (\textsc{STfe}) to derive meaningful features. The \textsc{STfe} processes the flattened mini-batch \(\mathbf{X'}\) and the corresponding holistic isolated graphs \(\{\mathcal{G}_i\}_{i=1}^B\), producing the output \(\mathcal{H} = \textsc{STfe}(\mathbf{X'}, \mathcal{G})\). Here, \(\mathcal{H} \in \mathbb{R}^{B' \times C_\text{out}}\) with \(C_\text{out}\) representing the output dimension. This extractor employs a deep residual architecture comprising multiple blocks, each containing multi-layer 1D dilated convolutions~\cite{dilated} to capture temporal dynamics across different receptive fields, and Chebyshev spectral graph convolutions~\cite{chebgcn} to extract intricate spatial features.
Detailed of the \textsc{STfe} can be found in Appendix~\ref{sec:stfe}.

As indicated in \textbf{Challenge 2}, the issue of data imbalance tends to push the \textsc{STfe} towards focusing on continual variables with abundant data, resulting in insufficient representations of data-limited expanding variables. Traditionally, over-sampling and data augmentation are employed to address such imbalances by generating more samples for the minority class. However, merely increasing the number of minority samples does not necessarily enrich their semantic information and may lead to overfitting. Additionally, augmenting time series data poses a particular challenge, as it requires introducing diversity without corrupting the inherent temporal patterns.
To circumvent these issues, we introduce a novel spatio-temporal focal learning approach. This method incorporates an innovative unsupervised \textsc{FocalCL} mechanism with negative pair filtering, specifically designed to adapt to the \textsc{FlatS} and enhance the feature extraction for expanding variables.

\subsubsection{Spatio-temporal Contrastive Learning}~\label{sec:STCL} 
% As discussed \textit{Challenge 2}, the imbalanced data pushes the model toward the well-trained continual variables, resulting in insufficient representations of expanding variables. 
% Traditionally, over-sampling and data augmentation are two common techniques used to address this issue by building more balanced samples. However, merely increasing the number of minority samples cannot enrich semantic information and may lead to overfitting. On the other hand, augmentation for time series data is particularly challenging because it is required to introduce diversity while preserving the temporal pattern without corruption.
Vanilla contrastive learning aims to learn temporal representations and capture variable heterogeneity by pulling positive samples closer together and pushing negative samples apart. It has shown superiority in robust representation learning when only limited data is available~\cite{liu2024timesurl, yue2022ts2vec}. The core components of contrastive learning include the construction of positive and negative sample pairs and the contrastive loss.

% To avoid the above problems, we turn to unsupervised time series contrastive learning~\cite{liu2024timesurl, yue2022ts2vec}, which learns contextual temporal representations and variable heterogeneity by attracting positive samples and separating negative samples, leading to more generalizable and robust representations. The vanilla contrastive learning includes positive and negative pair sample construction, the project layer, and contrastive loss.

Formally, to construct positive and negative sample pairs, a time series augmentation technique generates an additional view \(\mathbf{X'}^{\text{aug}} \sim p(\mathbf{X'}^{\text{aug}} | \mathbf{X})\). The augmentation conditional distribution \(p(\cdot | \mathbf{X})\) reflects the transformational impact of augmentation techniques on the temporal dynamics of the time series. A sample and its augmented counterpart are treated as a positive pair, i.e., $(\mathcal{X'}_i, \mathcal{X'}^{\text{aug}}_i)$, while the sample and the augmented counterparts of the other samples within the mini-batch are treated as negative pairs, i.e., \(\{(\mathcal{X'}_i, \mathcal{X'}^{\text{aug}}_j)\}_{j=1, j\neq i}^{B'}\). A projection layer maps the spatio-temporal features of the original $\mathcal{H}$ and augmented samples $\mathcal{H}^{\text{aug}}$ into latent space \(\mathcal{Z}\) and $\mathcal{Z}^{\text{aug}}$, respectively. The contrastive loss \(\mathcal{L}_\text{cl}\) for sample \(\mathcal{X'}_i\) is then formulated as:
\begin{gather}\label{clloss}
    \mathcal{L}_\text{cl}(\mathcal{X'}_i) = -log\frac{exp(s_{i,i}/\tau)}{\sum_{j=1} exp(s_{i,j}/\tau)},
\end{gather}
where \(s_{i,j}=sim(\mathcal{Z}_{i}, \mathcal{Z}^{\text{aug}}_{j})\) denotes the dot-product similarity measuring the feature similarity, 
% \(sim(u, v)\) denotes the dot product between \(u/||u||_2\) and \(v/||v||_2\), 
and \(\tau\) is the temperature factor that controls the magnitude of penalties on hard negative samples~\cite{wang2021understanding}. 

% Formally, a time series augmentation technique generates an additional view \(\mathbf{X}^{\text{(aug)}}\), which satisfies \(\mathbf{X}^{\text{(aug)}} \sim p(\mathbf{X}^{\text{(aug)}} | \mathbf{X})\). The augmentation conditional distribution \(p(\cdot | \mathbf{X})\) reflects the transformational impact of augmentation techniques on the temporal dynamics of the time series. Next, we treat the sample and its augmented sample of the variable at the same timestep as a positive pair, i.e., $(\mathbf{X}_i, \mathbf{X}^{\text{(aug)}}_i)$. The other augmented samples within a mini-batch are seen as negative pairs \(\{(\mathbf{X}_i, \mathbf{X}^{\text{(aug)}}_j)\}_{j=1, j\neq i}^{B}\). To measure the feature similarity in a latent space, the project layer with two linear transformers and the ReLU activation maps the two spatio-temporal representations \(\mathcal{H}^{(M)}\) and $\mathcal{H}^{(\text{aug}),(M)}$ into \(\mathcal{Z}\) and $\mathcal{Z}^{(\text{aug})}$. Finally, the contrastive loss \(\mathcal{L}_\text{cl}\) of sample \(\mathbf{X}_i\) is formulated as:
% \begin{gather}\label{clloss}
%     \mathcal{L}_\text{cl}(\mathbf{X}_i) = -log\frac{exp(s_{i,i}/\tau)}{\sum_{j=1} exp(s_{i,j}/\tau)},
% \end{gather}
% where \(s_{i,j}=sim(\mathcal{Z}_{i}, \mathcal{Z}^{(\text{aug})}_{j})\) measures the feature similarity, \(sim(u, v)\) denotes the dot product between \(u/||u||_2\) and \(v/||v||_2\), \(\tau\) is temperature factor used to control the magnitude of penalties on hard negative samples~\cite{wang2021understanding}. 

\subsubsection{Negatvie Pair Filter}~\label{sec:NF}
\begin{figure}[t]
    \centering
    \includegraphics[width=0.9\columnwidth]{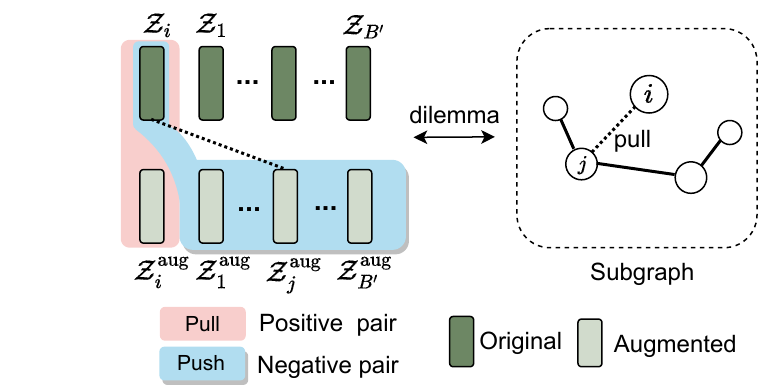}
    \caption{Contrastive samples with negative sample dilemma.}
    \label{fig:focal}
\vspace{-10pt}
\end{figure}
However, vanilla spatio-temporal contrastive learning introduces a unique dilemma regarding the treatment of negative sample pairs.
%Unlike vanilla contrastive learning, our flat scheme introduces a unique dilemma regarding the treatment of negative sample pairs. 
Specifically, as shown in Figure~\ref{fig:focal} left, a sample $\mathcal{X'}_i$ has its negative pair counterpart $\mathcal{X'}_j^\text{aug}$, where $j \neq i$. 
The learning process aims to push these samples apart, that is to make them dissimilar. 
However, a complication arises when $\mathcal{X'}_i$ and $\mathcal{X'}_j$ belong to the same subgraph $\mathcal{G}_i$ and are connected by an edge (Figure~\ref{fig:focal} right). In this case, $\mathcal{Z}_i$ and $\mathcal{Z}_j^\text{aug}$ should exhibit similar behaviours and thus should be pulled closer together during learning.
Experimentally, our ablation study supports this claim, where the performance decreases when vanilla contrastive learning is directly applied (refer to results in Figure~\ref{fig:abl}).
% a potential dilemma arises when the $i$-th node and its neighbors $(i+1)$-th and $(i-1)$-th nodes have their representations brought closer in the spatial module. 
% However, as negative samples in contrastive learning, their representations are pushed apart. 

In light of this, we propose to include a negative pair filter into vanilla contrastive learning that adaptively excludes negative sample pairs belonging to the same subgraph. Consequently, Equation~(\ref{clloss}) is re-formulated as:
\begin{gather}\label{negclloss}
    \mathcal{L}_\text{cl}(\mathcal{X'}_i) = -log\frac{exp(s_{i,i}/\tau)}{\sum_{j=1,j\notin \mathcal{G}_i}exp(s_{i,j}/\tau)},
\end{gather}
where \(\mathcal{G}_i\) denotes the subgraph containing sample $\mathcal{X'}_i$. 

\subsubsection{Focal Contrastive Learning (\textsc{FocalCL})}
When constructing negative pairs, we find that imbalanced data would lead to generating hard negative samples, which are excessively similar to positive samples. This makes it challenging for vanilla contrastive learning to learn discriminative representations.
To understand this issue formally, consider the following: let \((\mathbf{X'}_i, \mathbf{X'}^{\text{aug}}_j)\) be a negative pair, and \((\mathbf{X'}_i, \mathbf{X'}^{\text{aug}}_i)\) be a positive pair, where $\mathbf{X'}_i$ is an expanding variable sample and $\mathbf{X'}^{\text{aug}}_j$ is an augmented sample of a continual variable.
When the \textsc{STfe} is optimized with imbalanced data, it would focus on continual variables, overshadowing the distinct patterns of expanding variables. 
This results in an increase in the similarity between the negative pair embeddings \((\mathcal{Z}^{\text{aug}}_{i}, \mathcal{Z}^{\text{aug}}_{j})\), which diminishes the contrast between positive pair embeddings \((\mathcal{Z}^{\text{aug}}_{i}, \mathcal{Z}^{\text{aug}}_{i})\) and negative pair embeddings \((\mathcal{Z}^{\text{aug}}_{i}, \mathcal{Z}^{\text{aug}}_{j})\). Consequently, vanilla contrastive learning becomes less effective.
% \((sim(\mathcal{Z}^{\text{aug}}_{i}, \mathcal{Z}^{\text{aug}}_{j})\uparrow)\), thus increases the similarity between \(sim(\mathcal{Z}_{i}, \mathcal{Z}^{\text{aug}}_{i})\) and \(sim(\mathcal{Z}_{i}, \mathcal{Z}^{\text{aug}}_{j})\), making vanilla contrastive learning inefficient.

To alleviate this problem, our approach is to concentrate on the samples of expanding variables to compensate for the lack of optimization due to insufficient data. 
As pointed out by~\cite{wang2021understanding}, the temperature \(\tau\) in Equation (\ref{negclloss}) can control the penalty strength on the above hard negative samples. 
To improve the optimization of expanding variables, we further introduce a focal temperature factor. This factor adjusts the temperature \(\tau\) dynamically, thereby refining the learning process for these critical samples. The focal temperature factor can be formulated as follows:
\begin{gather}\label{eq:clnf}
    \tau_i =  \begin{cases} 
                \alpha * \tau, 0<\alpha<1.0, & \text{if } \mathbf{X'}_i \text{ is expanding variable,}  \\
                \tau & \text{others}. 
           \end{cases}
\end{gather}
%Based on Equation~\ref{negclloss}, The gradient concerning the negative similarity is formulated as:
% \begin{gather}
%     \frac{\partial \mathcal{L}(\mathcal{X'}_i)}{\partial s_{i,j}} = 
%     \frac{1}{\tau} {\frac{exp(s_{i,j}/\tau)}{\sum_{k, k\notin \mathcal{G}_i}exp(s_{i,k}/\tau)}}, \quad i \neq j.
% \end{gather}
% The imbalanced learning results in , and then the gradient magnitude increases.

% Although contrastive learning has the above hardness-aware property, it pays equal attention to the pair, where the anchor is a continual variable and the other is an expanding variable.

\subsubsection{Joint Objective Loss}
Finally, the output layer employs $1\times1$ convolutions to map the spatio-temporal features $\mathcal{H}$ to the desired $Q$ forecasting timesteps. The overall loss function integrates the focal contrastive loss (using Equation (\ref{eq:clnf}) as the temperature $\tau$ in Equation (\ref{negclloss}))with the mean absolute forecasting error loss. Formally, the joint loss is defined as:
\begin{gather}~\label{eq:final}
    \mathcal{L} = \frac{1}{B'}\sum_{i=1}^{B'}(-log\frac{exp(s_{i,i}/\tau_i)}{\sum_{j=1,j\notin \mathcal{G}_i}exp(s_{i,j}/\tau_i) }+|\mathbf{\hat{Y}}_i-\mathbf{Y}_i|).
\end{gather}
where $\mathbf{Y}=(\mathcal{Y}_1, \cdots, \mathcal{Y}_{B'})$ represents the ground truths.

\section{Experiments}~\label{sec:exp}
In this section, we comprehensively evaluate the proposed STEV model to investigate its forecasting performance, robustness to various expanding scenarios, and sensitivity to key parameters. Specifically, our experiments aim to answer the following research questions:
\begin{itemize}[leftmargin=*]
    \item \emph{RQ1:} How does STEV perform on real-world EVTS datasets compared to SOTA methods tailored to EVTSF? (see Section \ref{sec:overall})
    \item \emph{RQ2:} What is the impact of each component of STEV on its performance? (see Section \ref{sec:ablation})
    \item \emph{RQ3:} Does STEV have an advantage over other methods for addressing data imbalance, such as oversampling and augmentation? (see Section \ref{sec:ablation})
    \item \emph{RQ4:} How does STEV perform under different variable expanding scenarios? (see Section \ref{sec:explore})
\end{itemize}

\begin{table}[!ht]
\caption{Statistics of evaluation datasets. Notation \texttt{@} denotes different periods. The data split is train1/train2/valid/test.}
% \vspace{-10pt}
\label{tab:dataset}
\resizebox{\columnwidth}{!}{
\begin{tabular}{@{}ccccc@{}}
\toprule
Dataset & Sample Rate & \# Variables@$P_1$ & \# Variables@$P_2$ & Data Split     
\\ \midrule
EElectricity & 1 hour & 261 & 261+60 & 925d/7d/3d/321d \\
EPeMS  & 5 minutes& 296& 296+151& 63d/3d/2d/22d  \\
EWeather & 1 hour & 310 & 310+100& 560d/7d/3d/159d \\ \bottomrule
\end{tabular}}
% \vspace{-15pt}
\end{table}

%##################################################
\begin{table*}[!ht]
\caption{Overall EVTSF performance comparison between STEV and potential SOTA solutions. The \colorbox{customcolor}{bold values} indicate the best performance in the Oracle setting i.e., assuming complete data is available. The \textbf{bold} and \underline{value} formats indicate the best and suboptimal performance, respectively. The ‘-’ denotes \textit{None}, where TrafficStream is infeasible on the EElectricity and EWeather datasets due to the lack of a pre-defined graph.}
\label{tab:overall_results}
\resizebox{2\columnwidth}{!}{%
\begin{tabular}{@{}>{\centering\arraybackslash}p{0.1\linewidth}llcccccccccccc@{}}
\toprule
& \multicolumn{2}{l}{Dataset} & \multicolumn{4}{c}{EElectricity} & \multicolumn{4}{c}{EPeMS} & \multicolumn{4}{c}{EWeather} \\
\cmidrule(lr){4-7} \cmidrule(lr){8-11} \cmidrule(lr){12-15}
\multirow{-2}{*}{\begin{tabular}[c]{@{}l@{}}Training\\ Strategy\end{tabular}} & Model & Metric & Continual & Expanding & Overall & $\Delta$ & Continual & Expanding & Overall & $\Delta$ & Continual & Expanding & Overall & $\Delta$ \\ 
\cmidrule(lr){1-15}
% \midrule
 &  & MAE & {\cellcolor[HTML]{ECF4FF} \textbf{0.245}} & {\cellcolor[HTML]{ECF4FF} \textbf{0.200}} & {\cellcolor[HTML]{ECF4FF} \textbf{0.237}} & {-} & {\cellcolor[HTML]{ECF4FF} \textbf{21.97}} & {\cellcolor[HTML]{ECF4FF} \textbf{22.76}} & {\cellcolor[HTML]{ECF4FF} \textbf{22.23}} & {-} & {\cellcolor[HTML]{ECF4FF} 0.723} & {\cellcolor[HTML]{ECF4FF} 0.975} & {\cellcolor[HTML]{ECF4FF} 0.785} & - \\
 & \multirow{-2}{*}{AGCRN} & RMSE & {\cellcolor[HTML]{ECF4FF} \textbf{0.386}} & {\cellcolor[HTML]{ECF4FF} \textbf{0.316}} & {\cellcolor[HTML]{ECF4FF} \textbf{0.347}} & \textit{-} & {\cellcolor[HTML]{ECF4FF} 35.40} & {\cellcolor[HTML]{ECF4FF} \textbf{35.03}} & {\cellcolor[HTML]{ECF4FF} 35.28} & - & {\cellcolor[HTML]{ECF4FF} \textbf{1.197}} & {\cellcolor[HTML]{ECF4FF} \textbf{1.558}} & {\cellcolor[HTML]{ECF4FF} \textbf{1.294}} & - \\ \cmidrule(l){2-15} 
 &  & MAE & {\cellcolor[HTML]{ECF4FF}  0.281} & {\cellcolor[HTML]{ECF4FF}  0.235} & {\cellcolor[HTML]{ECF4FF}  0.273} & {-} & {\cellcolor[HTML]{ECF4FF}  22.67} & {\cellcolor[HTML]{ECF4FF}  23.55} & {\cellcolor[HTML]{ECF4FF}  22.97} & - & {\cellcolor[HTML]{ECF4FF}  \textbf{0.700}} & {\cellcolor[HTML]{ECF4FF}  \textbf{0.948}} & {\cellcolor[HTML]{ECF4FF}  \textbf{0.760}} & - \\
 & \multirow{-2}{*}{GWNET} & RMSE & {\cellcolor[HTML]{ECF4FF} 0.429} & {\cellcolor[HTML]{ECF4FF} 0.357} & {\cellcolor[HTML]{ECF4FF} 0.414} & {-} & {\cellcolor[HTML]{ECF4FF} \textbf{34.60}} & {\cellcolor[HTML]{ECF4FF} 35.09} & {\cellcolor[HTML]{ECF4FF}  \textbf{34.77}} & - & {\cellcolor[HTML]{ECF4FF} 1.199} & {\cellcolor[HTML]{ECF4FF} 1.590} & {\cellcolor[HTML]{ECF4FF} 1.305} & - \\ 
 \cmidrule(l){2-15} 
 &  & MAE & {\cellcolor[HTML]{ECF4FF} 0.338} & {\cellcolor[HTML]{ECF4FF} 0.339} & {\cellcolor[HTML]{ECF4FF} 0.338} & {-} & {\cellcolor[HTML]{ECF4FF} 25.26} & {\cellcolor[HTML]{ECF4FF} 26.46} & {\cellcolor[HTML]{ECF4FF} 25.67} & - & {\cellcolor[HTML]{ECF4FF} 0.961} & {\cellcolor[HTML]{ECF4FF} 1.321} & {\cellcolor[HTML]{ECF4FF} 1.049} & - \\
 & \multirow{-2}{*}{MSGNET} & RMSE & {\cellcolor[HTML]{ECF4FF} 0.485} & {\cellcolor[HTML]{ECF4FF} 0.468} & {\cellcolor[HTML]{ECF4FF} 0.482} & {-} & {\cellcolor[HTML]{ECF4FF} 39.02} & {\cellcolor[HTML]{ECF4FF} 39.47} & {\cellcolor[HTML]{ECF4FF} 39.17} & - & {\cellcolor[HTML]{ECF4FF} 1.559} & {\cellcolor[HTML]{ECF4FF} 2.096} & {\cellcolor[HTML]{ECF4FF} 1.705} & - \\ 
 \cmidrule(l){2-15} 
 &  & MAE & {\cellcolor[HTML]{ECF4FF} 0.374} & {\cellcolor[HTML]{ECF4FF} 0.345} & {\cellcolor[HTML]{ECF4FF} 0.369} & {-} & {\cellcolor[HTML]{ECF4FF} 25.35} & {\cellcolor[HTML]{ECF4FF} 25.48} & {\cellcolor[HTML]{ECF4FF} 25.39} & - & {\cellcolor[HTML]{ECF4FF} 0.830} & {\cellcolor[HTML]{ECF4FF} 1.102} & {\cellcolor[HTML]{ECF4FF} 0.896} & - \\
\multirow{-8}{*}{Oracle} & \multirow{-2}{*}{iTransformer} & RMSE & {\cellcolor[HTML]{ECF4FF} 0.581} & {\cellcolor[HTML]{ECF4FF} 0.541} & {\cellcolor[HTML]{ECF4FF} 0.574} & {-} & {\cellcolor[HTML]{ECF4FF} 40.48} & {\cellcolor[HTML]{ECF4FF} 40.42} & {\cellcolor[HTML]{ECF4FF} 40.46} & - & {\cellcolor[HTML]{ECF4FF} 1.450} & {\cellcolor[HTML]{ECF4FF} 1.868} & {\cellcolor[HTML]{ECF4FF} 1.562} & - \\ 
\midrule
 &  & MAE & 0.402 & 0.476 & 0.407 & 71.73\% & 26.91 & 27.33 & 27.05 & 21.68\% & 0.807 & 1.596 & 1.026 & 35.05\% \\
 & \multirow{-2}{*}{Nbeats} & RMSE & 0.620 & 0.856 & 0.641 & 84.73\% & 42.6 & 41.04 & 42.08 & 21.02\% & 1.416 & 2.580 & 1.816 & 40.37\% \\
 \cmidrule(l){2-15}
  &  & MAE & 0.538 & 0.507 & 0.532 & 124.47\% & 32.48 & 33.63 & 32.87 & 47.86\% & 0.919 & 1.186 & 0.984 & 29.86\% \\
 & \multirow{-2}{*}{PatchTST} & RMSE & 0.799 & 0.855 & 0.809 & 133.14\% & 50.55 & 50.10 & 50.40 & 45.92\% & 1.595 & 1.988 & 1.699 & 32.22\% \\
 \cmidrule(l){2-15}
  &  & MAE & 0.544 & 0.509 & 0.537 & 126.58\% & 32.02 & 33.26 & 32.44 & 42.07\% & 0.920 & 1.194 & 0.987 & 29.47\% \\
 & \multirow{-2}{*}{GPT4TS} & RMSE & 0.804 & 0.840 & 0.811 & 133.71\% & 49.38 & 49.45 & 49.40 & 44.95\% & 1.600 & 2.017 & 1.711 & 31.29\% \\
 \cmidrule(l){2-15}
 &  & MAE & 0.407 & 0.617 & 0.424 & 78.90\% & 27.41 & 27.90 & 27.58 & 24.07\% & 0.811 & 5.765 & 2.188 & 187.95\% \\
\multirow{-8}{*}{UTSF} & \multirow{-2}{*}{GRU} & RMSE & 0.620 & 1.319 & 0.700 & 101.73\% & 42.93 & 41.37 & 42.41 & 21.97\% & 1.414 & 9.830 & 5.321 & 311.20\% \\ \midrule
 &  & MAE & \textbf{0.248} & 0.643 & \underline{0.322} & \underline{35.86\%} & \underline{23.09} & 26.86 & \underline{24.37} & \underline{9.63\%} & \textbf{0.718} & 1.269 & 0.852 & 12.11\% \\
 & \multirow{-2}{*}{AGCRN} & RMSE & \textbf{0.388} & 1.438 & 0.712 & \underline{51.26\%} & 36.37 & 40.32 & 37.75 & 8.57\% & \textbf{1.220} & 2.122 & 1.491 & 15.22\% \\
  \cmidrule(l){2-15}
 &  & MAE & 0.307 & 0.526 & 0.348 & 46.84\% & 23.87 & \underline{25.94} & 24.57 & 10.53\% & 0.750 & \underline{1.160} & \underline{0.850} & \underline{11.84\%} \\
 & \multirow{-2}{*}{GWNET} & RMSE & 0.458 & 1.059 & 0.617 & 77.81\% & \underline{36.01} & \underline{38.41} & \underline{36.84} & \underline{5.95\%} & 1.284 & \underline{1.940} & \underline{1.471} & \underline{13.68\%} \\
  \cmidrule(l){2-15}
 &  & MAE & 0.350 & \underline{0.376} & 0.355 & 49.79\% & 26.46 & 31.79 & 28.26 & 27.13\% & 0.983 & 1.479 & 1.104 & 45.26\% \\
 & \multirow{-2}{*}{MSGNET} & RMSE & 0.505 & \underline{0.625} & \underline{0.530} & 52.74\% & 39.25 & 44.90 & 41.25 & 18.64\% & 1.591 & 2.308 & 1.793 & 38.56\% \\
   \cmidrule(l){2-15}
&  & MAE & 0.316 & 0.484 & 0.347 & 46.41\% & 31.50 & 33.90 & 32.31 & 45.34\% & 2.735 & 11.637 & 4.906 & 545.53\% \\
& \multirow{-2}{*}{Informer} & RMSE & 0.468 & 1.233 & 0.680 & 95.97\% & 50.10 & 51.74 & 50.70 & 45.82\% & 4.381 & 16.843 & 9.149 & 607.03\% \\
\cmidrule(l){2-15}
&  & MAE & 0.586 & 0.601 & 0.588 & 148.10\% & 32.39 & 33.99 & 32.93 & 48.13\% & 0.953 & 1.337 & 1.046 & 37.63\% \\
 & \multirow{-2}{*}{DLinear} & RMSE & 0.848 & 0.898 & 0.857 & 146.97\% & 49.91 & 50.39 & 50.07 & 44.00\% & 1.663 & 2.190 & 1.806 & 39.57\% \\
\cmidrule(l){2-15}
 &  & MAE & 0.382 & 0.413 & 0.388 & 63.71\% & 26.05 & 26.50 & 26.20 & 17.86\% & 0.830 & 1.172 & 0.896 & 17.89\% \\
\multirow{-12}{*}{FPTM} & \multirow{-2}{*}{iTransformer} & RMSE & 0.587 & 0.734 & 0.617 & 77.81\% & 41.15 & 41.47 & 41.26 & 18.67\% & 1.450 & 1.991 & 1.562 & 20.71\% \\ \midrule
 &  & MAE & - & - & - & - & 27.66 & 27.05 & 27.45 & 23.48\% & - & - & - & - \\
\multirow{-2}{*}{\begin{tabular}[c]{@{}l@{}}Continual \\ Learning\end{tabular}} & \multirow{-2}{*}{TrafficStream} & RMSE & - & - & - & - & 40.93 & 39.37 & 40.41 & 16.22\% & - & - & - & - \\ \midrule
\multicolumn{2}{l}{} & MAE & \underline{0.266} & \textbf{0.338} & \textbf{0.280} & \textbf{18.14\%} & \textbf{21.93} & \textbf{23.75} & \textbf{22.55} & \textbf{1.44\%} & \underline{0.727} & \textbf{1.067} & \textbf{0.810} & \textbf{6.58\%} \\
\multicolumn{2}{c}{\multirow{-2}{*}{STEV (ours)}} & RMSE & \underline{0.401} & \textbf{0.611} & \textbf{0.447} & \textbf{29.08\%} & \textbf{33.78} & \textbf{35.66} & \textbf{34.42} & \textbf{-1.01\%} & \underline{1.239} & \textbf{1.776} & \textbf{1.389} & \textbf{7.34\%} \\ \bottomrule
\end{tabular}}
% \vspace{-10pt}
\end{table*}
%##################################################

\subsection{Experimental setup}
\subsubsection{Datasets}
We used three real-world datasets across different domains for experiments: Electricity~\cite{mtgnn}, PeMS~\cite{liu2024largest}, Weather~\cite{rasp2020weatherbench}. 
As these datasets originally had full observations for all variables, we customized them to adapt to EVTSF. Table~\ref{tab:dataset} shows each dataset's statistics and data splits. 
For all datasets, both $H$ and $Q$ are set to 12. 
The details are as follows: \textbf{(i) EElectricity}: This dataset records hourly electricity consumption in kWh of 321 clients from 2012 to 2014. We randomly selected 261 clients as the continual variables over the pre-expanding period ($P_1$), and the remaining 60 clients as the expanding variables over the post-expanding period ($P_2$). \textbf{(ii) EPeMS}: The PeMS dataset consists of traffic flow data in District 7 of California. Since the sensor location is available, we first selected a central sensor as an anchor and then filtered 296 sensors as the continual variables, with the remaining 151 sensors as expanding variables, based on the spatial distance from the anchor. 
\textbf{(iii) EWeather:} Provided by WeatherBench, this dataset contains 410 nodes on the Earth sphere. We randomly selected 310 nodes as continual variables, and the remaining as expanding variables. In real-world applications, the expansion process varies across different contexts and requirements. Thus, We also tailored the datasets accordingly and explored various expanding scenarios (see Section~\ref{sec:explore}). More details can be found in Appendix~\ref{appendix:data}.

\subsubsection{Baselines}
We investigated and implemented a series of existing methods that can be tailored for EVTSF, classifying them into three types: \textbf{(1) UTS Forecasting (UTSF):} these methods forecast EVTS as individual UTS, naturally avoiding the inconsistent shape issue, including GRU~\cite{gru}, N-beats~\cite{NBeats}, PatchTST~\cite{PatchTST}, and GPT4TS~\cite{gpt4ts}; \textbf{(2) First-padding-then-masking MTSF (FPTM):} these methods first standardize data shapes through padding and then use SOTA MTSF models with a masking matrix to filter out padding loss, including GWNET~\cite{wu2019graph}, AGCRN~\cite{bai2020adaptive}, MSGNET~\cite{msgnet}, Informer~\cite{zhou2021informer}, DLinear~\cite{zeng2023transformers}, and iTransformer~\cite{liu2023itransformer}; \textbf{(3) Continual learning-based MTSF:} these methods train a model from scratch with train1 data ($P_1$) and then incrementally learn using train2 data ($P_2$), including TrafficStream~\cite{chen2021trafficstream}. 

Besides, we conducted an \textbf{Oracle} setting, assuming the availability of $P_1$ data for expanding variables, allowing each model to be optimized with fully observed data. While impractical, this setting helps verify the \textit{upper bound }for baselines benefiting from full observations. Further experimental implementations are given in Appendix~\ref{sec:ei}. 

\subsubsection{Evaluation metrics}
We evaluate forecasting performance using Mean Absolute Error (MAE) and Root Mean Square Error (RMSE) averaged over the forecasting horizon. Additionally, we propose to evaluate the performance gap to the \textit{upper bound }by quantifying the performance relative to the Oracle as 
\begin{gather}
    \Delta = \frac{E_\textit{Expanding} - E_\textit{Oracle}}{E_\textit{Oracle}}.
\end{gather}
% A lower $\Delta$ indicates better performance and a negative $\Delta$ implies that the expanding setting outperforms the Oracle setting.

\subsection{Overall Results}\label{sec:overall}
Table~\ref{tab:overall_results} reports the overall performance of different methods on three datasets, including the averaged MAE and RMSE on continual, expanding, and overall variables over 12 timesteps (\emph{RQ1}). 
Note that TrafficStream relies on a pre-defined graph to transfer learned information from continual to expanding variables, therefore, we only report the performance on the EPeMS dataset. 
We observe the following findings: (1) STEV achieves SOTA performance across almost all settings, showing significant average improvements with average reductions of 10.72\% in MAE and 11.85\% in RMSE compared to the runner-up.
This success is primarily attributed to the innovative spatio-temporal focal learning, which effectively focuses on expanding variables. However, STEV does not show improvements for the continual variables in the electricity and weather datasets. This may be due to the coarse-grained temporal modeling with CNNs, akin to the observed performance differences between GWNET and AGCRN.
(2) Due to the scarcity of data, the performance of expanding variables declines significantly across all methods. While the GPT4TS and PatchTST separately benefit from the pre-trained large model and weight-sharing mechanism for few-shot learning, they lack explicit spatio-temporal patterns, leading to inferior performance. In contrast, STEV, using only 5\% of the complete data (3-day EPeMS data), surpasses methods trained with the Oracle setting, demonstrating its ability to effectively learn representations for expanding variables.
(3) Although the FPTM-based strategy introduces padding values, it captures spatio-temporal correlations more effectively than the solely temporal modeling used in UTSF-based strategies, leading to better overall performance.
(4) The continual learning approach benefits from leveraging a pre-trained model and incrementally learning from new observations during the expansion period. However, this can lead to catastrophic forgetting, which negatively affects the model’s performance on both continual and expanding variables.
% \vspace{-12pt}
\subsection{Ablation Study}\label{sec:ablation}
We perform an ablation study on the EPeMS data to verify the effectiveness of each component of STEV (\emph{RQ2}). We review four components of STEV, including 
(1) \emph{\textsc{FlatS}} which allows STEV to learn spatio-temporal dependencies under varying variables; 
(2) \emph{Contrastive Learning} (CL) which enhances STEV to learn robust representations for both continual and expanding variables;
(3) \emph{Negative Filter} (NF) which allows STEV to exclude inappropriate negative pairs in contrastive learning; 
(4) \emph{\textsc{FocalCL}}  which enables STEV to focus on expanding variables and thus alleviate the impact of imbalanced data.

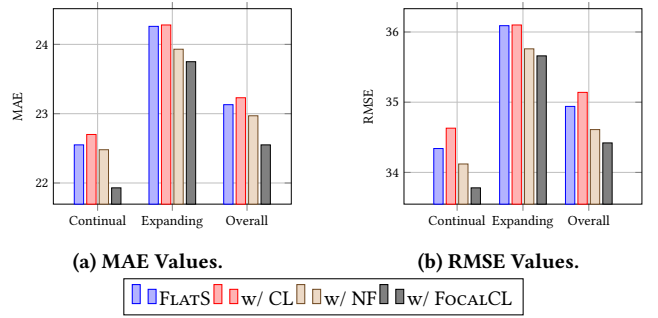
\begin{figure}[!t] % 使用 figure* 环境以跨越两栏
    \begin{subfigure}[b]{0.45\columnwidth} % 使用0.48\textwidth确保两个子图填充整个双栏
        \centering
        \resizebox{\linewidth}{!}{
        \begin{tikzpicture}
            \begin{axis}[
                ybar,
                bar width=6pt, % 调整柱状体宽度
                symbolic x coords={Continual, Expanding, Overall},
                xtick=data,
                ylabel={MAE},
                ylabel style={yshift=-12pt}, % 调整纵坐标名称与坐标轴之间的距离
                height=6cm,
                enlarge x limits={abs=1.0cm}, % 调整组之间的间隔
                grid=major,
                ]
                \addplot coordinates {(Continual,22.55) (Expanding,24.26) (Overall,23.13)};
                \addplot coordinates {(Continual,22.70) (Expanding,24.28) (Overall,23.23)};
                \addplot coordinates {(Continual,22.48) (Expanding,23.93) (Overall,22.97)};
                \addplot coordinates {(Continual,21.93) (Expanding,23.75) (Overall,22.55)};
            \end{axis}
        \end{tikzpicture}}
        \caption{MAE Values.}
        \label{fig:mae}
    \end{subfigure}
    \hfill
    \begin{subfigure}[b]{0.45\columnwidth} 
        \centering
        \resizebox{\linewidth}{!}{
        \begin{tikzpicture}
            \begin{axis}[
                ybar,
                bar width=6pt, % 调整柱状体宽度
                symbolic x coords={Continual, Expanding, Overall},
                xtick=data,
                ylabel={RMSE},
                ylabel style={yshift=-12pt}, % 调整纵坐标名称与坐标轴之间的距离
                height=6cm,
                enlarge x limits={abs=1.2cm}, % 调整组之间的间隔
                grid=major,
                legend entries = {\textsc{FlatS}, w/ CL, w/ NF, w/ \textsc{FocalCL}},
                legend columns = -1,   
                legend style={nodes={scale=0.55, transform shape},font=\huge},
                legend to name={legendperformance1},
                legend pos= north west,
                ]
                \addplot coordinates {(Continual,34.34) (Expanding,36.09) (Overall,34.94)};
                \addplot coordinates {(Continual,34.63) (Expanding,36.10) (Overall,35.14)};
                \addplot coordinates {(Continual,34.12) (Expanding,35.76) (Overall,34.61)};
                \addplot coordinates {(Continual,33.78) (Expanding,35.66) (Overall,34.42)};
            \end{axis}
        \end{tikzpicture}}
        \caption{RMSE Values.}
        \label{fig:rmse}
    \end{subfigure}
    \ref{legendperformance1}
    \caption{Ablation Study of STEV in terms of MAE and RMSE.}
    \label{fig:abl}
% \vspace{-10pt}
\end{figure}

We gradually add each of the above modules starting from \textsc{FlatS}. Note that when \textsc{FocalCL} is added, it forms STEV.
From Fig.~\ref{fig:abl} we observe a continuous performance improvement, except for the variant w/ CL. 
As discussed in Section~\ref{sec:NF}, 
there is a dilemma between the naive CL and spatial module. We thus propose the negative filter to remove conflict negative samples.

We additionally compared two conventional solutions for data imbalance (\textit{RQ3}), i.e., oversampling (\emph{OS}) and data augmentation (\emph{DA}). 
In our experiments, the double sampling rate and the mixup augmentation technique~\cite{mixup} are utilized on the expanding data. 
Table~\ref{tab:abl} shows that the performance is comparable with data-level techniques, and using oversampling even leads to a decrease. To explore the effect of the ultra data augmentation, we conduct the \emph{DA*} experiment, augmenting the entire dataset. Despite the improved performance, the computational overhead doubled.

\begin{table}[!htbp]
\caption{Comparison of various conventional solutions for the imbalanced issue on EPeMs.}
\vspace{-10pt}
\label{tab:abl}
\resizebox{0.9\columnwidth}{!}{
\begin{tabular}{@{}lcccccc@{}}
\toprule
\multirow{2}{*}{Variants} & \multicolumn{2}{c}{Continual} & \multicolumn{2}{c}{Expanding} & \multicolumn{2}{c}{Overall} \\ 
\cmidrule(lr){2-3} \cmidrule(lr){4-5} \cmidrule(lr){6-7}
 & MAE & RMSE & MAE & RMSE & MAE & RMSE \\ \midrule
STEV & \textbf{21.93} & \textbf{33.78} & \textbf{23.75} & \textbf{35.66} & \textbf{22.55} & \textbf{34.42} \\
\textsc{FlatS} & 22.55 & 34.34 & 24.26 & 36.09 & 23.13&34.94\\
\textsc{FlatS} w/ OS & 22.88 & 34.41 & 24.49 & 36.06 & 23.42 & 34.98 \\
\textsc{FlatS} w/ DA & 22.43 & 34.19 & 24.36 & 36.02 & 23.16 & 34.83 \\ 
\textsc{FlatS} w/ DA* & 22.19 & 33.95 & 24.11 & 36.09 & 22.84 & 34.69 \\
\bottomrule
\end{tabular}}
% \vspace{-14pt}
\end{table}

\begin{figure*}[!htbp]
    \centering
    % 第一张子图
    \begin{subfigure}[b]{0.3\textwidth}
        \centering
        \includegraphics[width=\textwidth]{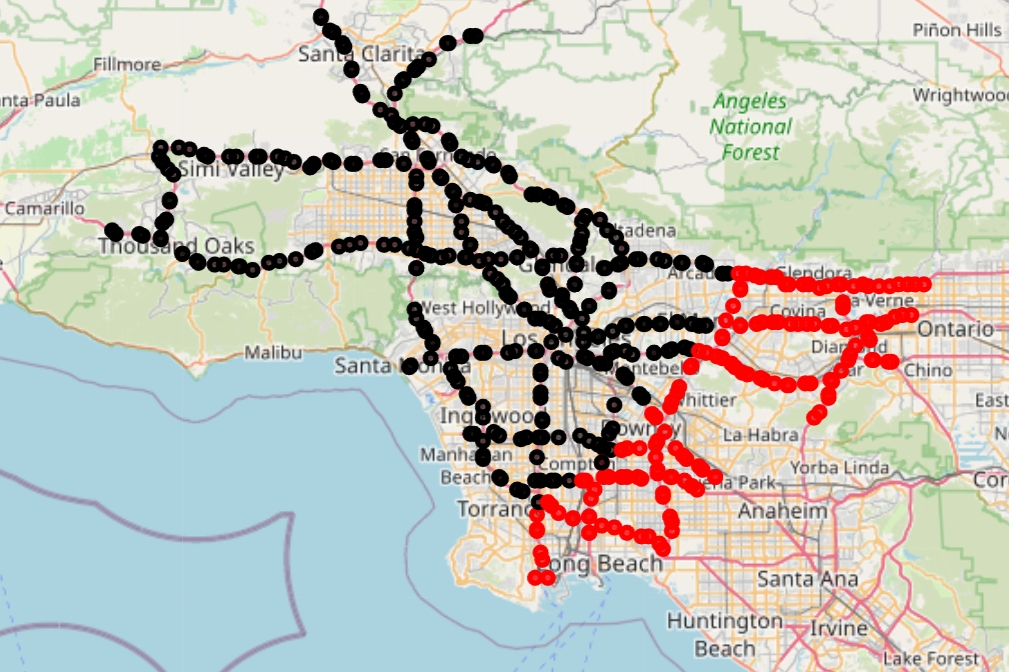}
        \caption{Area Expansion.}
        \label{fig:area}
    \end{subfigure}
    \hfill
    \begin{subfigure}[b]{0.3\textwidth}
        \centering
        \includegraphics[width=\textwidth]{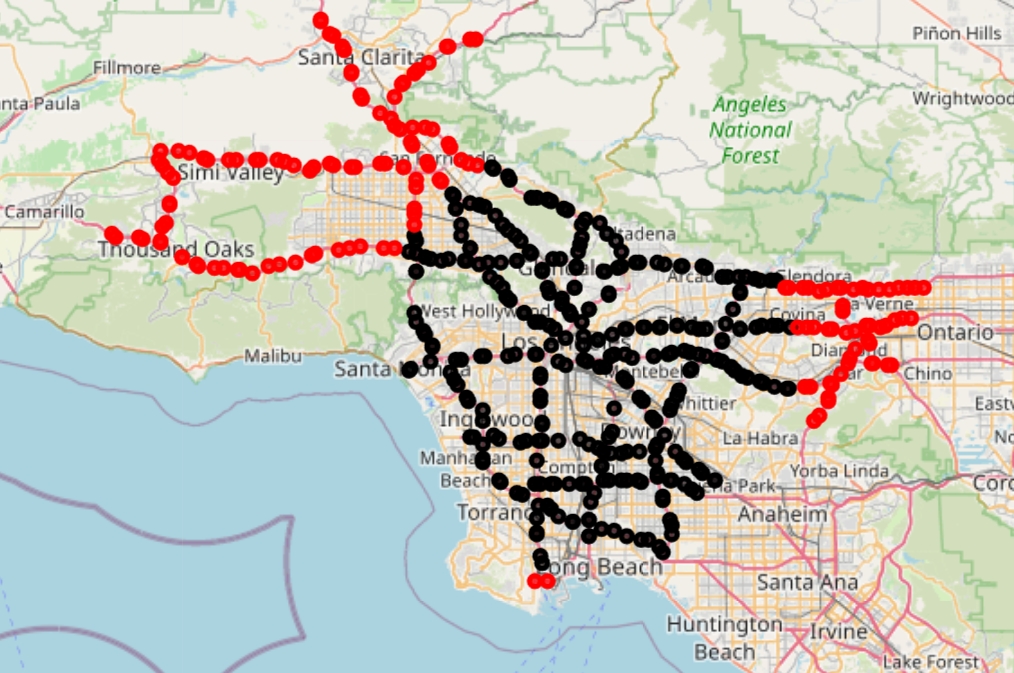}
        \caption{Spatial Expansion.}
        \label{fig:spatial}
    \end{subfigure}
    \hfill
    \begin{subfigure}[b]{0.3\textwidth}
        \centering
        \includegraphics[width=\textwidth]{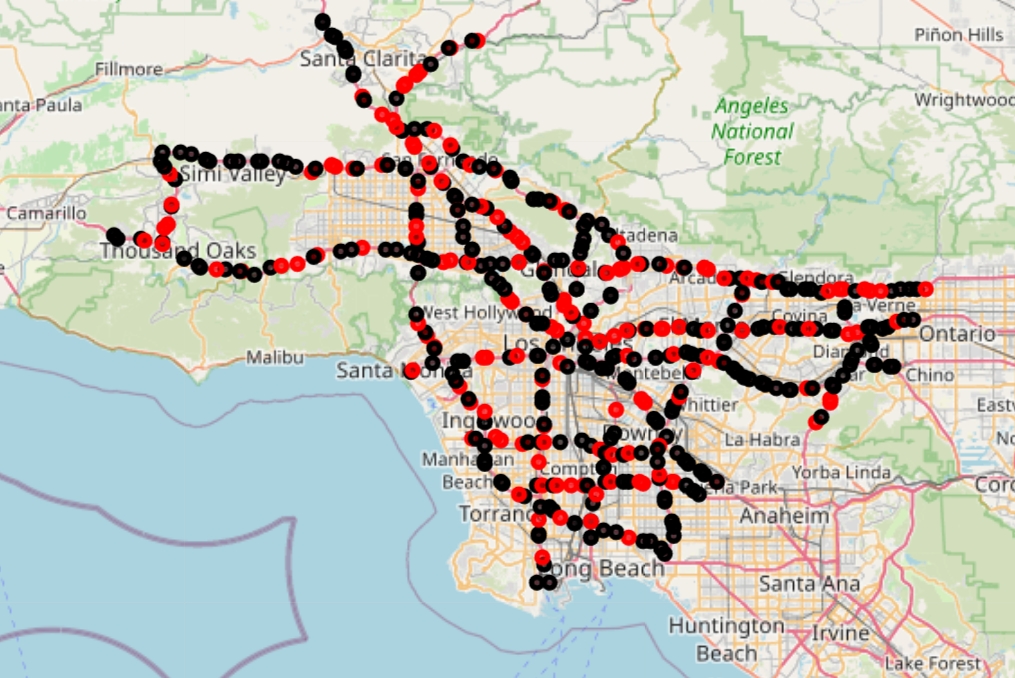}
        \caption{Internal Expansion.}
        \label{fig:random}
    \end{subfigure}
    \begin{subfigure}[b]{0.3\textwidth}
        \centering
        \includegraphics[width=\textwidth]{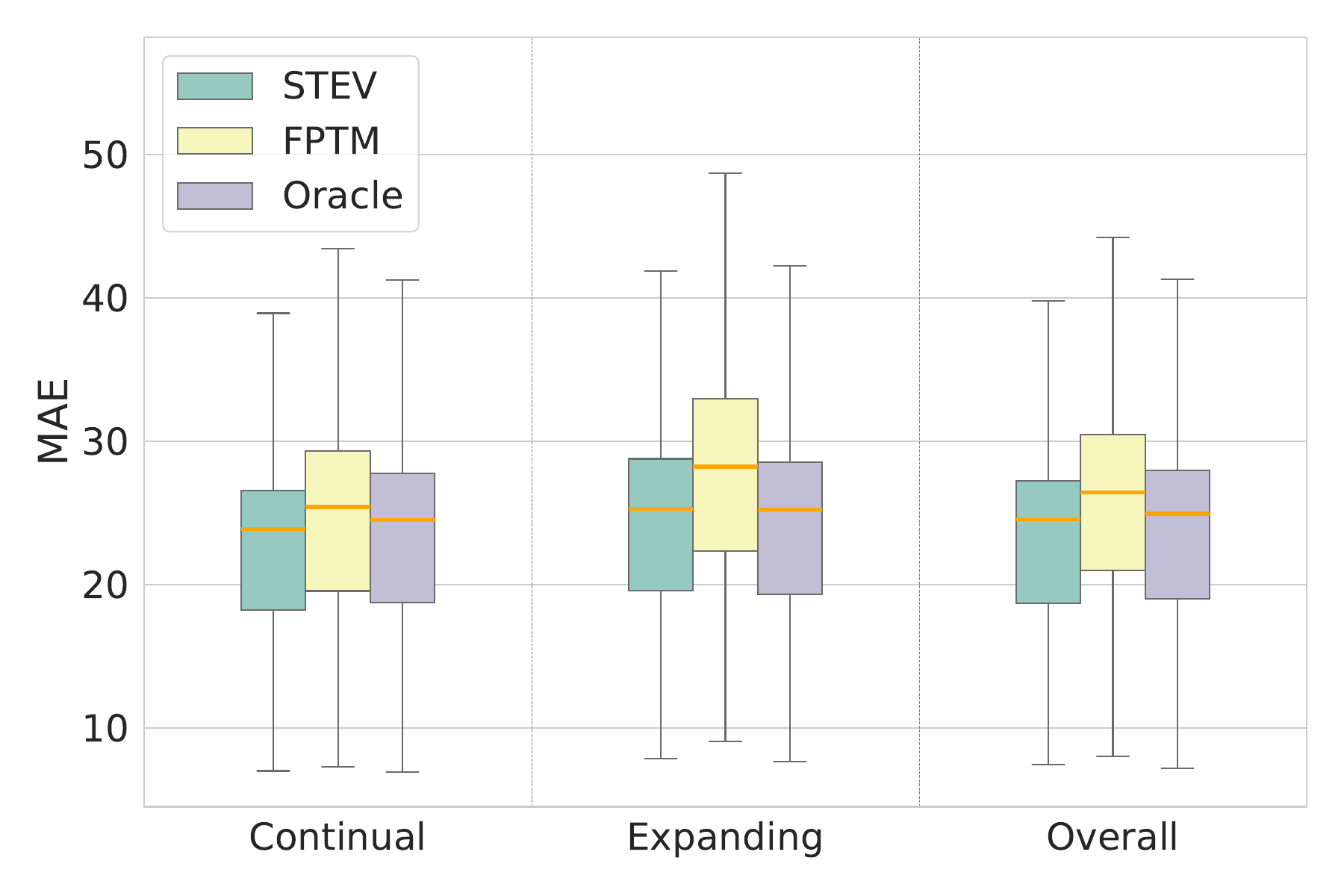}
        \caption{Performance with Area Expansion.}
        \label{fig:parea}
    \end{subfigure}
    \hfill
    \begin{subfigure}[b]{0.3\textwidth}
        \centering
        \includegraphics[width=\textwidth]{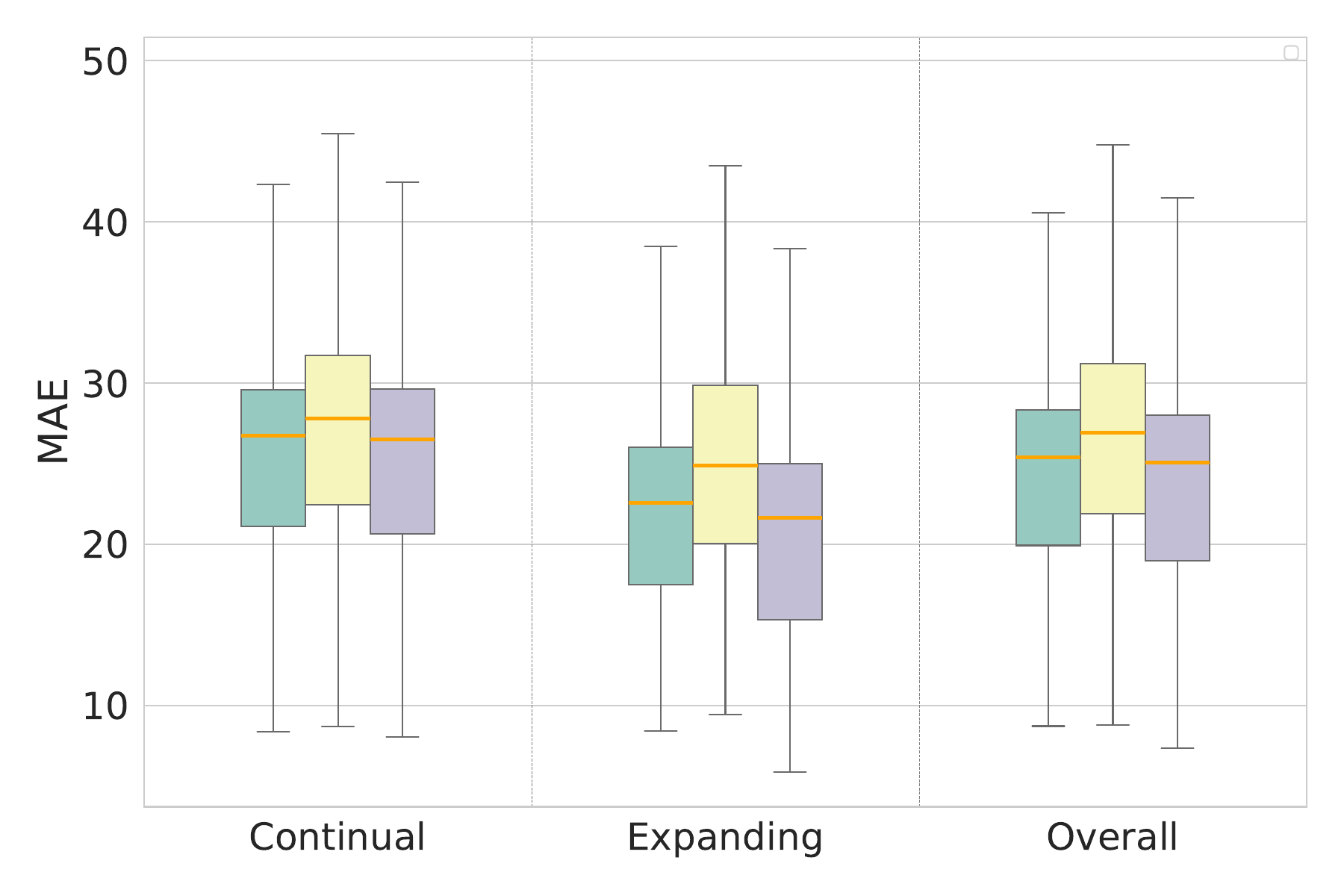}
        \caption{Performance with Spatial Expansion.}
        \label{fig:pspatial}
    \end{subfigure}
    \hfill
    \begin{subfigure}[b]{0.3\textwidth}
        \centering
        \includegraphics[width=\textwidth]{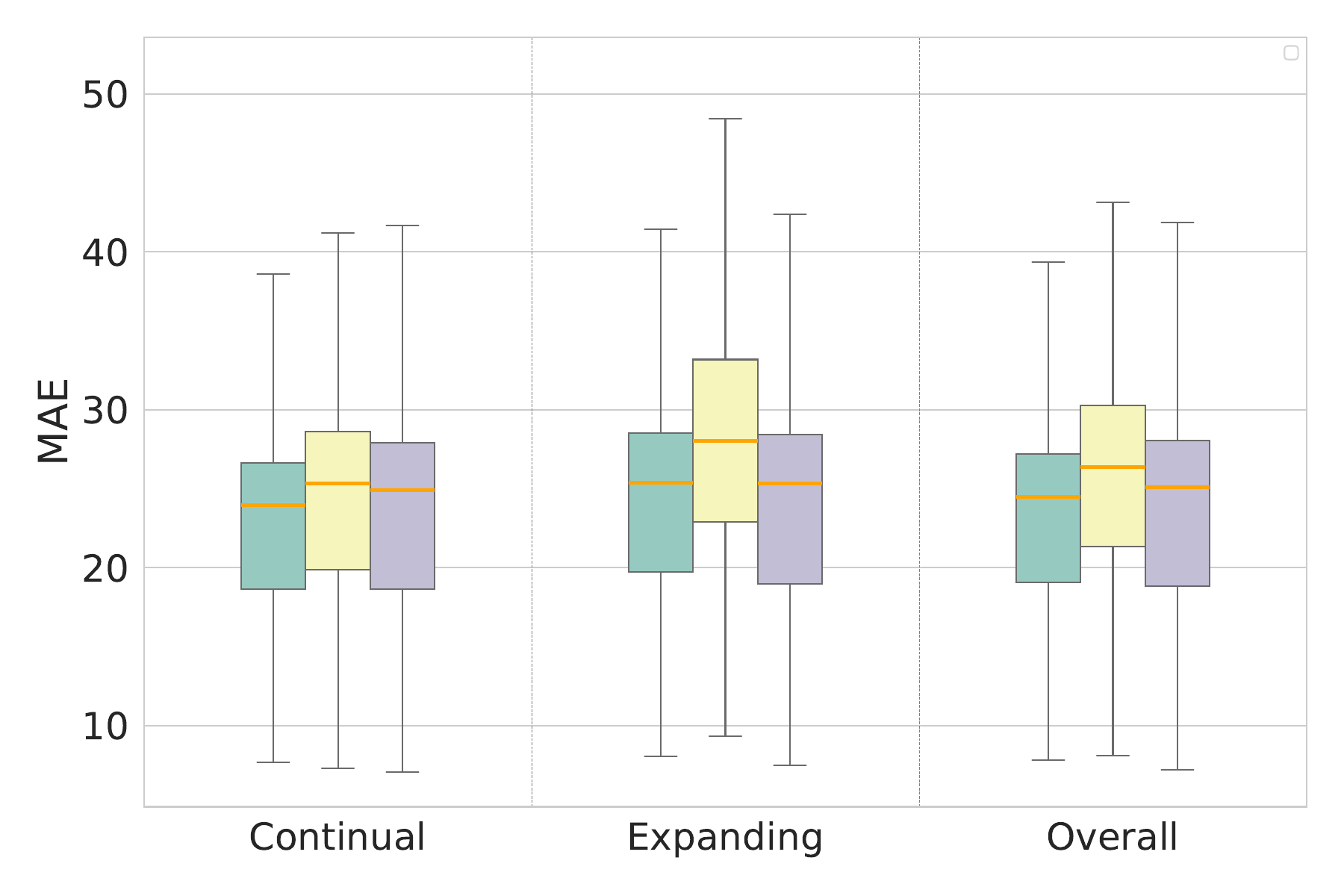}
        \caption{Performance with Internal Expansion.}
        \label{fig:prandom}
    \end{subfigure}
    \caption{Visualization of geographic distribution (a)-(c) and performance comparison with three expanding scenarios (d)-(f). In (a)-(c), \textbf{black nodes} represent continual variables, and \textcolor{red}{red} nodes indicate expanding variables.}
    \label{fig:expanding}
% \vspace{-10pt}
\end{figure*}

\begin{table}[!ht]
\centering
\caption{Consecutive expansion setting}
% \vspace{-10pt}
\label{tab:sec}
\resizebox{\columnwidth}{!}{%
\begin{tabular}{cccccc}
\hline
\textbf{\#Var@P1} & \textbf{\#Var@P2} & \textbf{\#Var@P3} & \textbf{\#days@train1} & \textbf{@train2} & \textbf{@train3} \\ \hline
296 & 427 & 447 & 59 & 4 & 3 \\ \hline
\end{tabular}%
}
\vspace{-12pt}
\end{table}

\begin{table}[!ht]
\centering
\caption{Performance on the consecutive Expansion.}
% \vspace{-10pt}
\label{tab:per_sec}
\resizebox{\columnwidth}{!}{
\begin{tabular}{cccccc}
\hline
\multicolumn{2}{c}{Model} & Con@P1 & Exp@P2 & Exp@P3 & Overall \\ \hline
\multirow{2}{*}{STEV} & MAE & 22.40 & 23.21 & 23.78 & 22.74 \\
 & RMSE & 34.19 & 34.79 & 37.94 & 34.49 \\ \hline
\multirow{2}{*}{FPTM-GWNET} & MAE & 23.25 & 24.11 & 25.55 & 23.62 \\
 & RMSE & 35.25 & 35.93 & 38.55 & 35.64 \\ \hline
\end{tabular}%
}
% \vspace{-12pt}
\end{table}

\begin{table}[htbp]
\centering
\caption{Performance on a longer observation period.}
% \vspace{-10pt}
\label{tab:longer}
\resizebox{0.9\columnwidth}{!}{%
\begin{tabular}{llccc}
\hline
\multicolumn{2}{l}{\textbf{Model}} & \textbf{Continual (52d)} & \textbf{Expanding (14d)} & \textbf{Overall} \\ \hline
\multirow{2}{*}{STEV} & MAE & 22.17 & 23.05 & 22.47 \\
 & RMSE & 33.95 & 34.52 & 34.15 \\ \hline
\multirow{2}{*}{FPTM-GWNET} & MAE & 22.70 & 24.62 & 23.35 \\
 & RMSE & 34.67 & 36.69 & 35.37 \\ \hline
\end{tabular}%
}
\end{table}
\subsection{Expanding Scenario Exploration}~\label{sec:explore}
To explore the generalizability of STEV in real-world scenarios, we conduct comprehensive experiments on the EPeMS dataset under various expansion settings, involving the partition of continual and expanding variables, the number of expansions, and the duration of the expansion period (RQ4). 

We first simulate three potential demands of expanding variables in the traffic domain, including (1) Area-based expansions where new sensors are deployed in specific focus areas, with those in the top-left region remaining as continual variables (see Fig.~\ref{fig:area}); (2) Spatial-based expansions where sensors spread to the surrounding areas with the expansion of urban monitor planning. A central sensor and 296 sensors within a 30 km radius serve as continual variables, while others are expanding variables (see Fig.~\ref{fig:spatial}); 
and (3) Internal expansions where the denser sensor network increases perception density. We randomly select 296 sensors as continual variables (see Fig.~\ref{fig:random}).
As shown in Fig.~\ref{fig:expanding}, STEV consistently outperforms FPTM-GWNET~\cite{wu2019graph} across all three scenarios, demonstrating strong generalization capability.

To assess the adaptability, we conduct a consecutive experiment involving two-time spatial expansions. Table~\ref{tab:sec} details the variable statistics and durations, while Table~\ref{tab:per_sec} reports performance results. The findings show that STEV consistently outperforms FPTM-GWNET across all metrics, for both continual and expanding variables, confirming its robustness in handling dynamic spatial expansions.

We further examine STEV’s performance over the much longer period of sensor deployment by conducting experiments on the expanding period over 14 days.
Table~\ref{tab:longer} shows that STEV maintains superiority over FPTM-GWNET, even as data scarcity diminishes over time. Compared to the 3-day results (Table~\ref{tab:overall_results}), STEV effectively captures evolving spatiotemporal patterns, demonstrating its stability across both short- and long-term deployments.

\subsection{Long-term Expanding-variate Time Series Forecasting}
To validate the effectiveness of STEV in long-term EVTSF, we conduct a 48-step forecasting experiment with 48-step inputs. 
Apart from the extended input and output lengths, all experimental settings are kept identical to those in the 12-step forecasting.
Table~\ref{tab:long-term} shows that STEV substantially outperforms both GWNET and iTransformer, across all evaluation metrics. The consistent performance gain demonstrates STEV's generalization capability and its practical value for real-world time series forecasting tasks that require long-range forecasting accuracy.

\begin{table}[htbp]
\centering
\caption{Performance on a long-term EVTSF.}
\resizebox{0.8\columnwidth}{!}{
\begin{tabular}{@{}ccccc@{}}
\toprule
Method & Metric & Continual & Expanding & AVG. \\ \midrule
\multirow{2}{*}{FPTM-GWNET} & MAE & 36.45 & 40.11 & 37.68 \\
 & RMSE & 53.98 & 57.55 & 55.21 \\ \midrule
\multirow{2}{*}{FPTM-iTransformer} & MAE & 43.12 & 41.57 & 42.60 \\ \cmidrule(l){2-5} 
 & RMSE & 68.01 & 65.39 & 67.14 \\ \midrule
\multirow{2}{*}{STEV} & MAE & \textbf{29.47} & \textbf{31.26} & \textbf{30.07} \\
 & RMSE & \textbf{44.93} & \textbf{46.01} & \textbf{45.29} \\ \bottomrule
\end{tabular}
}
\label{tab:long-term}  
\end{table}

\subsection{Average Forgetting Evaluation}
To further validate the effectiveness of the retraining paradigm, we compare the performance of TrafficSteam --- a continual learning-based method with two retraining-based approaches, GWNET (the champion of baselines) and STEV, on the EPeMS dataset. The comparison is conducted in terms of average forecasting MAE (AFMAE)~\cite{chaudhry2018efficient}, which reflects the extent of knowledge forgetting and can be formulated by $AFMAE = MAE\_NEW - MAE\_OLD$. We first introduce the two notions as follows:
\begin{itemize}
    \item MAE\_OLD@427 indicates the model trained on data \textit{before expansion}, then evaluated in terms of MAE on ``old tasks'' (427 sensors) on the test set.
    \item MAE\_NEW@427 indicates the model trained on data \textit{after expansion}, then evaluated the in terms of MAE on ``old tasks'' (427 sensors) on the test set.
\end{itemize}
A lower AFMAE value corresponds to a lower degree of forgetting. As shown in Table~\ref{tab:afmae}, the retraining-based methods retain previously learned knowledge, whereas the continual learning method suffers from knowledge forgetting.

\begin{table}[h]
\caption{Average forgetting MAE.}
\resizebox{0.8\columnwidth}{!}{
\begin{tabular}{@{}cccc@{}}
\toprule
Method & AFMAE & MAE\_OLD@427 & MAE\_NEW@427 \\ \midrule
FPTM-GWNET & -1.52 & 24.09 & 22.56 \\
STEV & -0.22 & 22.79 & 22.57 \\
TrafficStream & 1.03 & 26.63 & 27.66 \\ \bottomrule
\end{tabular}
}
\label{tab:afmae}
\end{table}

\section{Related Work}
\textbf{Multivariate Time Series Forecasting.} Extensive deep learning-based MTSF methods prove that both capturing intra-series temporal dependencies and inter-series spatial dependencies are beneficial for modeling future changes~\cite{zhang2017deep, shi2015convolutional, liu2023itransformer, NEURIPS2023_dc1e32dd, lai2024lightcts, liu2025timekd}. Particularly, spatio-temporal graph neural networks such as Graph WaveNet~\cite{wu2019graph} and AGCRN~\cite{bai2020adaptive} further enhance spatio-temporal representations by combining GNNs with LSTM and GNNs with TCN, respectively.
However, these works cannot be directly performed in the EVTSF task. Even with padding strategies, their performance still suffers from data imbalance. Recent efforts~\cite{chen2021trafficstream, wang2023pattern} have explored the expansion of traffic networks where observed traffic areas are continuously expanding as new sensors are deployed. 
These studies primarily address this problem as a two-phase learning task within the continual learning or online learning paradigm~\cite{chen2018lifelong}. 
The main challenge that these studies intend to address, is the catastrophic forgetting phenomenon~\cite{french1999catastrophic}. They first discover the continual variables that need to learn new patterns, then employ historical data replay~\cite{rolnick2019experience}, elastic weight consolidation~\cite{kirkpatrick2017overcoming}, and memory bank mechanism~\cite{lopez2017gradient} to consolidate historical knowledge. In contrast, the performance achieved by the retraining paradigm is an upper bound of continual learning due to the global optimization for model parameters with better convergence and performance~\cite{kirkpatrick2017overcoming, rolnick2019experience}. Another similar task to EVTSF is spatio-temporal Kriging~\cite{wu2021inductive}, a.k.a. spatio-temporal extrapolation~\cite{hu2023graph}, which performs imputation for expanding variables based on the context of continual variables. Although both are aimed at expanding variables, it has different learning objectives from EVTSF.

\noindent{\textbf{Contrastive Learning in Time Series.}} Recently, unsupervised contrastive learning has gained significant traction in the time series domain~\cite{yue2022ts2vec, tonekaboni2021unsupervised, liu2024timesurl}. The primary objective of this approach is to make positive samples attractive and separate negative samples, thereby enabling the learning of inherent temporal characteristics. Prior efforts mainly explore the effect of positive and negative pairs. For instance, motivated by the local smoothness of a signal during the generative process, TNC~\cite{tonekaboni2021unsupervised} encourages the consistency of samples within a temporal neighborhood. TS2Vec~\cite{yue2022ts2vec} learns contextual information between timesteps at different temporal resolutions. In this approach, two augmented views of the same time step exhibit similarity and are considered dissimilar otherwise. To further explore the impact of augmentation, TimesURL~\cite{liu2024timesurl} employs a frequency-temporal-based augmentation technique to encourage temporal consistency. However, due to the imbalanced data and the potential conflict between negative samples in contrastive learning and neighbors in the graph structure, directly leveraging contrastive learning is hard to benefit EVTSF (refer to Table~\ref{tab:abl}).

\section{Conclusion}
In this paper, we introduced a new task called Expanding Multivariate Time Series Forecasting (EVTSF), driven by the practical needs of sensing expansion in Cyber-Physical Systems. To tackle this, we propose STEV, a flexible spatio-temporal forecasting framework to address two key challenges: inconsistent data shape and imbalanced spatio-temporal learning. Specifically, our approach features a Flat Scheme that unifies data shapes with a simple flattening operation, making the model agnostic to variable scale. Besides, we employed Focal Contrastive Learning to enhance the learning of robust and discriminative spatio-temporal feature representations, with a focal contrastive loss that emphasizes optimizing expanding variables. We conducted extensive experiments on three real-world datasets, demonstrating that STEV significantly enhances performance on variables with limited observations and achieves results comparable to, or better than, methods using complete data. Furthermore, an intriguing avenue for future research could involve exploring the out-of-distribution problem, particularly how data distribution shifts after the deployment of new sensors.

\begin{acks}
This work was supported by the National Natural Science Foundation of China (Nos. U2468207, 62272398, 62402420) and Sichuan Science and Technology Program (Nos. 2024NSFTD0036, 2024NSFJQ0019)
\end{acks}
% \balance
\bibliographystyle{ACM-Reference-Format}
\bibliography{main}

\newpage
\appendix
\section{Reproducibility}
This section outlines the details of STEV implementation to ensure reproducibility.

\subsection{Details of Spatio-temporal Feature Extractor}~\label{sec:stfe}

\begin{figure}[ht]
    \centering
    \includegraphics[width=0.8\columnwidth]{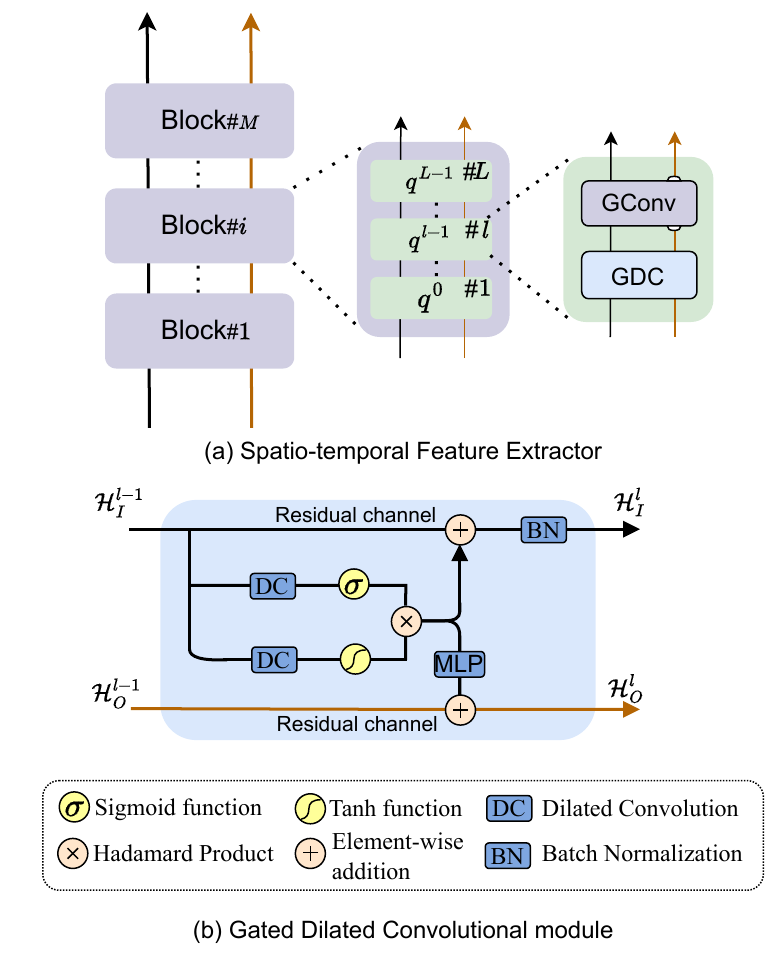}
    \caption{The architecture of Spatio-temporal Feature Extractor.}
    \label{fig:stfe}
\end{figure}

To efficiently capture multi-scale temporal dependencies and spatial dependencies, following \cite{mtgnn}, the spatio-temporal feature extractor employs a deep residual architecture that incorporates 
Gated Dilated Convolutional module (GDC) with 1D dilated convolution~\cite{dilated} and Graph Convolutional module (GConv) with chebyshev spectral graph convolution~\cite{chebgcn}.
As depicted in Figure~\ref{fig:stfe} (a), the basic block contains a multi-layer spatio-temporal convolutional architecture with various receptive fields. 
For the sake of simplicity, we focus on describing the operation of the $l$-th layer of the $i$-th block in detail. 
Given the hidden features $\mathcal{H}_I^{l-1}$, and $\mathcal{H}_O^{l-1}$, representing the inner and outer residual connections, and $\mathcal{H}_I^{0}=\text{Embed}(\mathbf{X}), \mathcal{H}_O^{0}=\mathbf{0}$, then the two outputs of $l$-th layer features can be formulated as,
\begin{gather}
    \mathcal{U} = tanh(\mathcal{H}_I^{(l-1)} \star f^d_{1\times k}) \odot \sigma(\mathcal{H}_I^{(l-1)} \star f^d_{1\times k}) \\
    \mathcal{H}_I^{(l)} = \text{BatchNorm}(\mathcal{H}_I^{(l-1)} + \mathcal{U}) \\
    \mathcal{H}_O^{(l)} = \text{MLP}(\mathcal{H}_O^{(l-1)}) + \mathcal{U}\\
    \mathcal{H}_I^{(l)} = \sum_{j=1}^{J} \mathbf{C}^{(j)} \cdot \mathbf{\Theta}^{(j)}
\end{gather}
where $\mathcal{H} \star f_{1\times k}$ denotes dilated convolution~\cite{wavenet} with kernel size $k$ and dilated factor $d=q^{l-1}$ and $q$ is an increased rate; for spectral graph convolution, $\mathbf{C}^{(1)} = \mathcal{H}_I^{(l)}$, $\mathbf{C}^{(2)} = \mathbf{\hat{L}}\mathcal{H}_I^{(l)}$, $\mathbf{C}^{(j)} = 2 \cdot \mathbf{\hat{L}} \mathbf{C}^{(j-1)} - \mathbf{C}^{(j-2)}$, $\mathbf{\hat{L}}$ represents the symmetric normalization $\mathbf{L} = \mathbf{I} - \mathbf{D}^{-1/2} \mathbf{A} \mathbf{D}^{-1/2}$, $\mathbf{D}$ is the degree matrix, and $\mathbf{\Theta}\in \mathbb{R}^{C_{in}\times C_{out}}$ is a parameter matrix.

\begin{figure}[!ht]
    \centering
    \includegraphics[width=0.8\columnwidth]{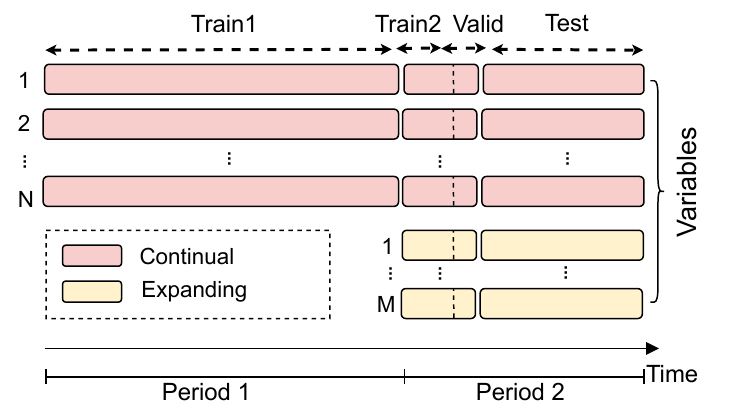}
    \caption{Illustration of Data Partition.}
    \label{fig:dataset}
\vspace{-14pt}
\end{figure}

\subsection{Data}~\label{appendix:data}
We craft three real-world datasets across different domains for the EVTSF experiments, including Electricity~\cite{mtgnn}, PeMS~\cite{liu2024largest}, and WeatherBench~\cite{rasp2020weatherbench}, due to the lack of public-availability EVTS data. Originally, these data have complete observations for all variables, we therefore customer them to adapt to the EVTSF task. Fig.~\ref{fig:dataset} visualizes the partition of training, valid, and test sets. Notably, there are two partitions of training data: train1 and train2, representing training data before and after expansion, respectively. For variable partitioning, we first determine quantities based on empirical proportions, then utilize random selection, and spatial center to filter variables depending on whether location is available. As for the duration of data collection for the expanding period, a minimum periodic observation is adopted based on the principle of timely prediction. Table~\ref{tab:dataset} shows the statistics and data splits for each dataset. In real-world applications, the expansion process varies across different contexts and requirements. Thus, we also tailored the datasets accordingly and explored various expanding scenarios (see Section~\ref{sec:explore}). The details of each dataset are as follows:
\begin{itemize}
    \item \textbf{EElectricity}: This dataset records hourly electricity consumption in kWh of 321 clients from 2012 to 2014. In this context, expansion refers to the new clients integrated into the grid. Due to the absence of location information and without loss of generality, we randomly selected 261 clients as the continual variables over the pre-expanding period ($P_1$), and the remaining 60 clients as the expanding variables over the post-expanding period ($P_2$).
    \item \textbf{EPeMS}: The PeMS dataset involves sensors monitoring traffic flow in District 7 of California, where variables represent sensors. In this context, expansion refers to the installation of new traffic flow sensors at different locations. Since the sensor location is available, we first selected a central sensor as an anchor and then filtered 296 sensors as the continual variables based on the spatial distance from the anchor, with the remaining 151 sensors as expanding variables.
    \item \textbf{EWeather:} The dataset records the temperature from 410 stations on the Earth sphere. Similar to EElectricity, We randomly selected 310 stations as continual variables, and the remaining as expanding variables.
\end{itemize}

Figure~\ref{fig:dataset} further visualizes the partition of training, valid, and test sets. Note that there are two partitions of training data: train1 and train2, representing training data before and after expansion, respectively.

\subsection{Baselines}
We investigated and implemented a series of existing methods that can be tailored for EVTSF. They are classified into three types: 
\begin{itemize}
    \item \textbf{UTS Forecasting (UTSF):} These methods treat the EVTSF task as the individual UTS forecasting, naturally avoiding the inconsistent shape issue, including GRU~\cite{gru}, a recurrent neural network variant to capture temporal dependencies by leveraging gating mechanisms; Nbeats~\cite{NBeats}, a deep neural architecture that combines the backward and forward residual connections, and deep stacked fully connected layers; PatchTST~\cite{PatchTST}, a Transformer with subseries-level patches and channel-independence; and GPT4TS~\cite{gpt4ts}, a fine-tuning time series analysis architecture by pre-trained large model.
    
    \item \textbf{First-padding-then-masking MTSF (FPTM):} We standardize data shape through padding operation and then employ SOTA MTSF models. We also utilize a masking matrix to filter out padding loss to alleviate the data noise impact caused by the filling value. The SOTA methods include GWNET~\cite{wu2019graph}, a spatio-temporal graph convolutional network with a self-learning adjacency matrix and stacked dilated 1D convolution to capture spatio-temporal dependencies; AGCRN~\cite{bai2020adaptive}, an adaptive graph convolutional recurrent network that employs graph convolution gated recurrent units to capture interdependencies among variables and intra-variable temporal dependencies; MSGNET~\cite{msgnet}, a spatio-temporal graph network that learns the inter-series correlations across multiple time scales with frequency domain analysis; and iTransformer~\cite{liu2023itransformer}, a Transformer that captures multivariate correlations via attention mechanism and the feed-forward network on variate tokens.
    \item \textbf{Continual learning-based MTSF:} These methods train a model from scratch with the initial data ($P_1$) and then incrementally learn using the expanding data ($P_2$). The TrafficStream~\cite{chen2021trafficstream} is included that consolidates the knowledge learned previously and transfers it to the current model to learn traffic expansion and evolving patterns.
\end{itemize}

\subsection{Experimental Implements}~\label{sec:ei}
All methods are conducted with NVIDIA GeForce RTX 3090Ti with PyTorch library. The Adam with a learning rate of 1e-3 is to optimize parameters. The early stopping is used to confirm the optimal weights in which the training process ends until the performance on the valid set does not improve within 10 patients.  For all datasets, the historical observations over 12 time steps are inputted to obtain the predictions over the next 12 time steps, i.e., both $H$ and $Q$ are set to 12. The focal factor $\alpha$ is set to 0.3 for three datasets. Apart from the factor, we follow the hyper-parameters setting with MTGNN~\cite{mtgnn}. Specifically, the batch size is set to 16. The number of blocks $M$ is set to 4. The number of layers $L$ is set to 2. The dimensions of node and time embeddings are set to 20 and 10. In the gated dilated convolutional module, the dilated rate $q$ is set to 2, and the kernel size is set to 2. The temporal and spatial convolution modules both have 32 output channels. 

\section{Hyperparameter Study}~\label{sec:para_study}
\begin{table}[!hpth]
\caption{Varying $\alpha$ of focal contrastive loss.}
\label{tab:para}
\begin{tabular}{c|cccccc}
\toprule
\multirow{2}{*}{$\alpha$}& \multicolumn{2}{c}{Continual} & \multicolumn{2}{c}{Expanding} & \multicolumn{2}{c}{Overall} \\ 
\cmidrule(lr){2-3} \cmidrule(lr){4-5} \cmidrule(lr){6-7}
 & MAE & RMSE & MAE & RMSE & MAE & RMSE \\ 
\cmidrule(lr){1-1}\cmidrule(lr){2-7}
0.05 & 22.10 & 33.88 & 23.80 & 35.72 & 22.68 & 34.51 \\
0.1 & 21.96 & 33.71 & 23.59 & 35.43 & 22.51 & 34.30 \\
0.3 & 21.93 & 33.78 & 23.75 & 35.66 & 22.55 & 34.42 \\
0.5 & 22.13 & 34.02 & 23.80 & 35.60 & 22.69 & 34.56 \\
0.7 & 22.20 & 33.98 & 23.86 & 35.46 & 22.76 & 34.49 \\
1.0 & 22.48 & 34.12 & 23.93 & 35.56 & 22.97 & 34.61\\
\bottomrule
\end{tabular}%
\end{table}

The STEV model introduces a novel hyperparameter, the focal weight $\alpha$, which plays a crucial role in controlling the strength of attention directed toward expanding variables. 
Table~\ref{tab:para} presents the experimental results under various settings of $\alpha$, highlighting its impact on model performance.
Notably, when $\alpha = 1.0$, our loss function aligns with the commonly used contrastive loss, ensuring a balanced treatment of both continual and expanding variables.
As $\alpha$ decreases, the optimization process increasingly prioritizes the representation learning of expanding variables, thereby improving their performance. However, this shift comes at a potential cost: excessive emphasis on expanding variables can lead to over-optimization, which may adversely affect the performance of the continual variables. These findings emphasize the importance of carefully tuning $\alpha$ to achieve a balanced trade-off between continual and expanding variable modeling, ensuring optimal performance across the dataset.

\section{Evaluation on Different Data Augmentations}
To explore the impact of data augmentation on model performance, we evaluate several augmentation strategies, including Mixup, Jitter, and the hybrid approach combining drift and quantization. As shown in Table~\ref{tab:augmentation}, the hybrid method achieves the best overall performance, indicating its effectiveness in enhancing model generalization of contrastive learning.

\begin{table}[htbp]
\centering
\caption{Performance on different data augmentations}
% \resizebox{0.6\columnwidth}{!}{%
\begin{tabular}{@{}ccccc@{}}
\toprule
Methods &  & Con. & Exp. & AVG. \\ \midrule
\multirow{2}{*}{Hybrid} & MAE & 21.93 & 23.75 & 22.55 \\
 & RMSE & 33.78 & 35.66 & 34.42 \\ \midrule
\multirow{2}{*}{Mixup~\cite{mixup}} & MAE & 23.16 & 24.88 & 23.74 \\
 & RMSE & 35.08 & 36.63 & 35.61 \\ \midrule
\multirow{2}{*}{Jitter} & MAE & 22.34 & 24.07 & 22.92 \\
 & RMSE & 34.02 & 35.96 & 34.69 \\ \bottomrule
\end{tabular}%
% }
\label{tab:augmentation}
\end{table}

\begin{table}[htbp]
\centering
\caption{Computational and memory analysis}
\label{tab:computation}
\resizebox{0.8\columnwidth}{!}{%
\begin{tabular}{ccc}
\hline
Model & MFLOPs & Parameters(M) \\ \hline
STEV & 522.494 & 0.7252 \\
GWNET & 843.049 & 0.3082 \\
PatchTST & 33835.253 & 6.3472 \\
iTransformer & 5652.025 & 6.3257 \\ \hline
\end{tabular}%
}
\end{table}
\begin{figure*}[!ht]
    \centering
    % 第一张子图
    \begin{subfigure}[b]{0.8\textwidth}
        \centering
        \includegraphics[width=\textwidth]{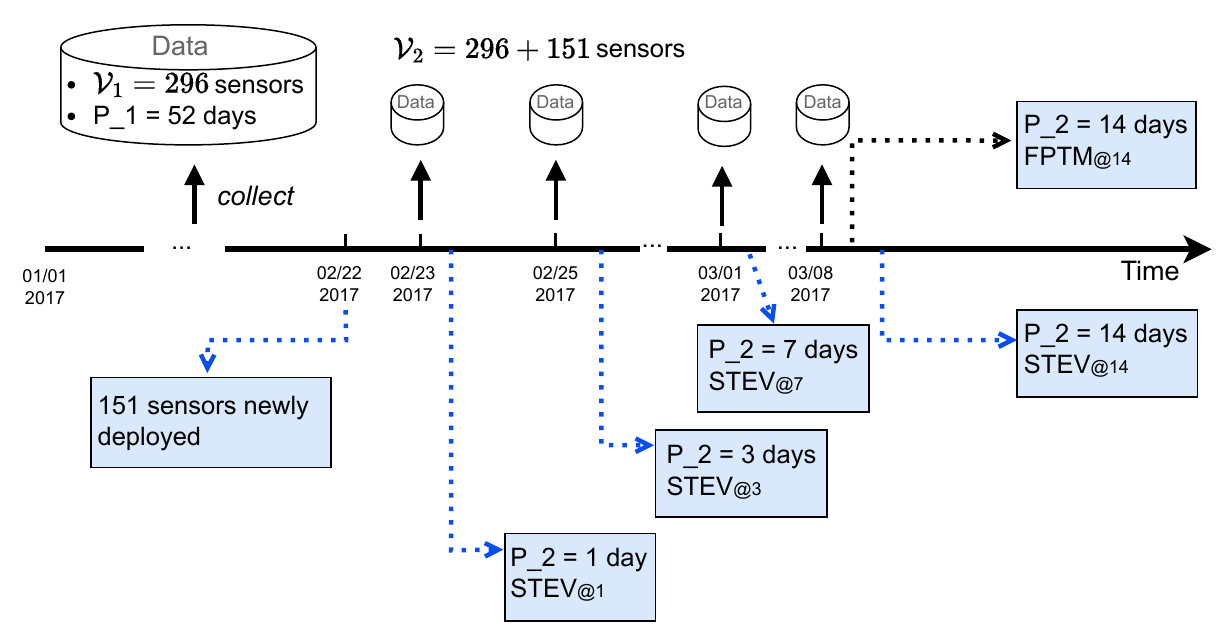}
        \caption{The traffic flow forecasting model training procedure at various waiting times for data collection. $\mathcal{V}_1$ and $\mathcal{V}_2$ denote two variable sets of the before and after expansion. $P_1$ and $P_2$ denote two periods before and after expansion.}
        \label{fig:case_example}
    \end{subfigure}
    \vfill
    \begin{subfigure}[b]{0.8\columnwidth}
        \includegraphics[width=\textwidth]{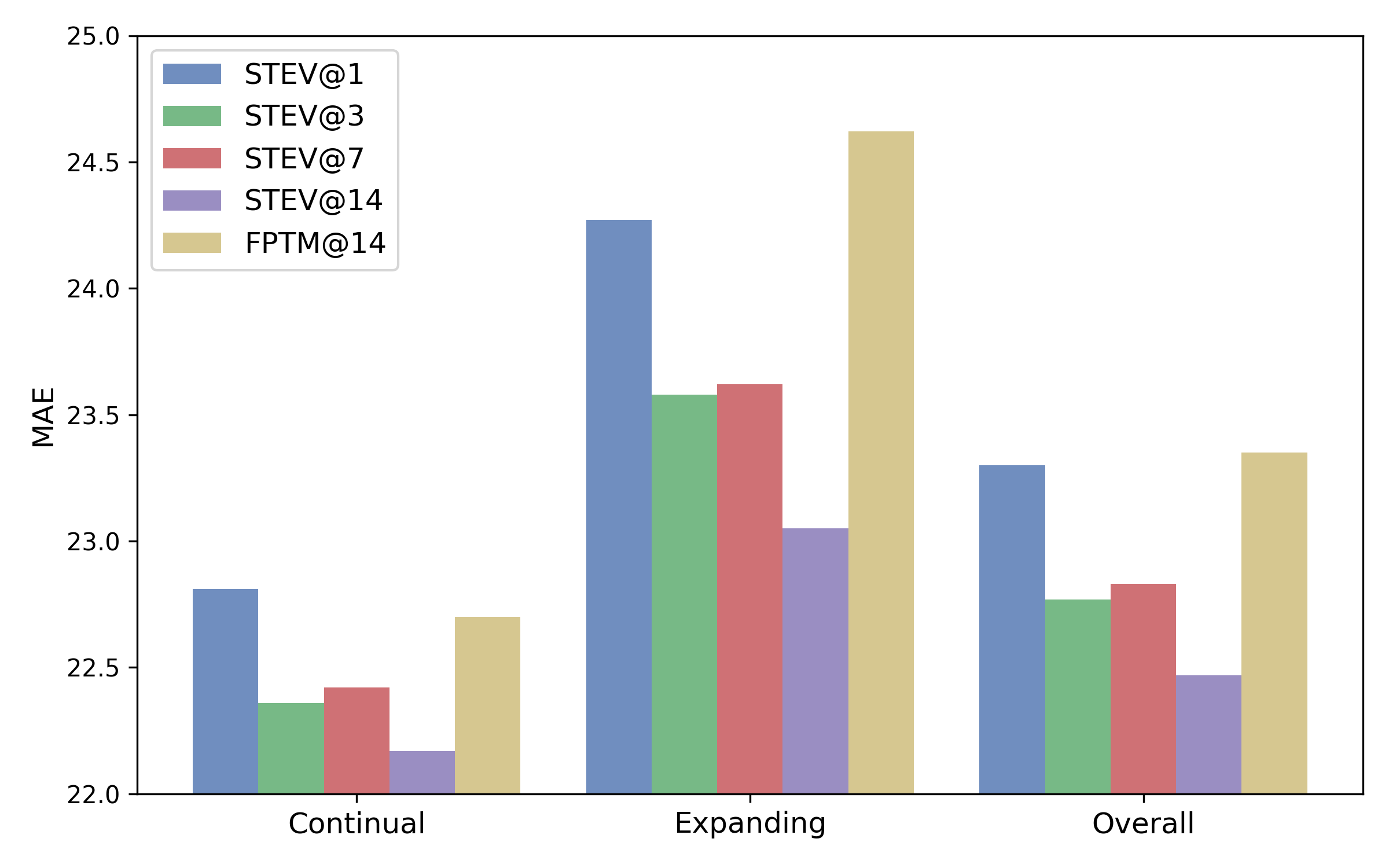}
        \caption{Performance on MAE.}
        \label{fig:case_mae}
    \end{subfigure}
    % \hfill
    \begin{subfigure}[b]{0.8\columnwidth}
        % \centering
        \includegraphics[width=\textwidth]{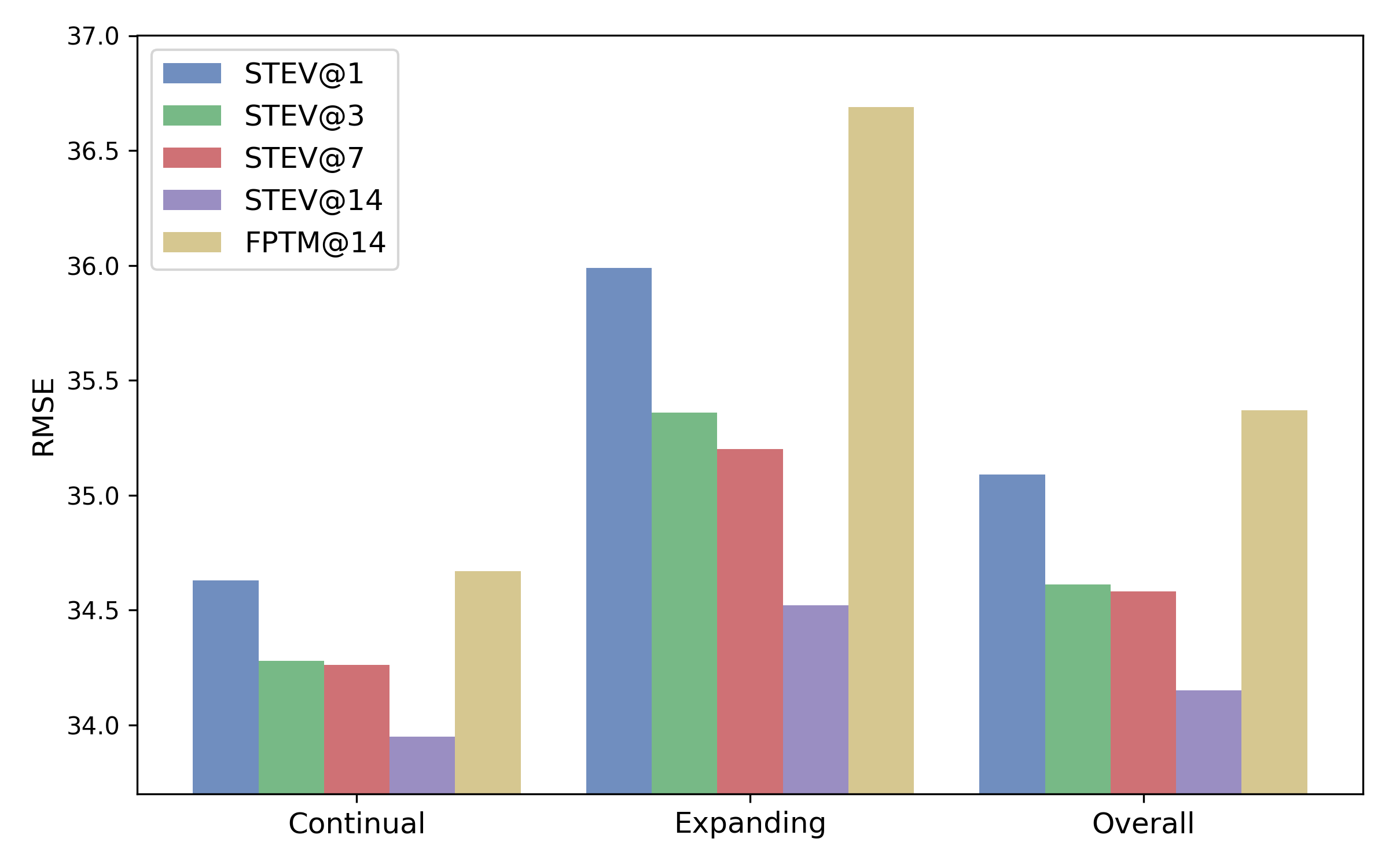}
        \caption{Performance on RMSE.}
        \label{fig:case_rmse}
    \end{subfigure}
    \caption{Case Study}
    \label{fig:case}
\end{figure*}
\section{Appendix: Empirical Study}
\subsection{Efficiency Analysis}
Table~\ref{tab:computation} provides a comparative analysis of the computational and memory efficiency of the proposed STEV model against several baseline methods, focusing on the number of floating-point operations (MFLOPs) and the total number of model parameters. The results show that STEV strikes a remarkable balance between model size and computational efficiency, enabling a highly feasible solution for real-world applications.

\subsection{Case Study}
We further investigate the effectiveness of STEV in timely prediction through a case study. Figure~\ref{fig:case_example} presents the model learning process at various waiting times for data collection. In this study, data for the first period ($P_1$) were collected over 52 days from 296 sensors, after which 151 additional sensors were deployed to expand the sensing range for traffic flow monitoring in District 7, California. In this context, the forecasting model is keen to be trained after collecting considerable data. To simulate different data collection scenarios, we conducted model training using four distinct durations representing waiting times after the deployment of new sensors: 1 day, 3 days, 7 days, and 14 days. For comparison, we also utilized the runner-up baseline method, FPTM-GWNET, trained with a total of 66 days of data ($P_1+P_2=52+14$). Fig.~\ref{fig:case_mae} and Fig.~\ref{fig:case_rmse} report the forecasting performance on MAE and RMSE. 

The results reveal that STEV, even when trained with just 1-day data, outperforms FPTM-GWNET trained on the 14-day data. This demonstrates the strong capability of STEV to quickly adapt to newly added variables and deliver accurate forecasts, meeting the requirements of the proposed expanding-variate time series forecasting task for timely predictions. Furthermore, the performance evaluations across the four durations show that STEV not only effectively learns spatio-temporal dependencies under extreme data imbalance (with only 1 day of data) but also achieves consistent improvements as the volume of augmented variable data increases (up to 14 days). These findings highlight the adaptability of STEV in handling dynamic and evolving spatio-temporal data for real-world timely prediction tasks.
\end{document}